\documentclass{article}

\usepackage{PRIMEarxiv}

\usepackage[utf8]{inputenc} 
\usepackage[T1]{fontenc}    
\usepackage{hyperref}       
\usepackage{url}            
\usepackage{booktabs}       
\usepackage[table]{xcolor} 
\usepackage{multirow}      
\usepackage{amsfonts}       
\usepackage{nicefrac}       
\usepackage{microtype}      
\usepackage{lipsum}
\usepackage{fancyhdr}       
\usepackage{graphicx}       
\graphicspath{{media/}}     

\title{Evaluating Large Language Models with Grid-Based Game Competitions: An Extensible LLM Benchmark and Leaderboard
}

\author{
  Oguzhan Topsakal, Colby J. Edell, Jackson B. Harper \\
  Computer Science Department \\
  Florida Polytechnic University \\
  Lakeland, Florida, 33805\\
  \texttt{otopsakal@floridapoly.edu}
  }

\begin{document}
\maketitle

\begin{abstract}
We introduce a novel and extensible benchmark for large language models (LLMs) through grid-based games such as Tic-Tac-Toe, Connect Four, and Gomoku. The open-source game simulation code, available on GitHub, allows LLMs to compete and generates detailed data files in JSON, CSV, TXT, and PNG formats for leaderboard rankings and further analysis. We present the results of games among leading LLMs, including Claude 3.5 Sonnet and Claude 3 Sonnet by Anthropic, Gemini 1.5 Pro and Gemini 1.5 Flash by Google, GPT-4 Turbo and GPT-4o by OpenAI, and Llama3-70B by Meta. We also encourage submissions of results from other LLMs. In total, we simulated 2,310 matches (5 sessions for each pair among 7 LLMs and a random player) across three types of games, using three distinct prompt types: list, illustration, and image. The results revealed significant variations in LLM performance across different games and prompt types, with analysis covering win and disqualification rates, missed opportunity analysis, and invalid move analysis. The details of the leaderboard and result matrix data are available as open-access data on GitHub. This study enhances our understanding of LLMs' capabilities in playing games they were not specifically trained for, helping to assess their rule comprehension and strategic thinking. On the path to Artificial General Intelligence (AGI), this study lays the groundwork for future exploration into their utility in complex decision-making scenarios, illuminating their strategic thinking abilities and offering directions for further inquiry into the limits of LLMs within game-based frameworks.
\end{abstract}

\keywords{large language model 
\and LLM 
\and benchmark
\and evaluate
\and performance
\and test
\and leaderboard
\and competition
\and championship
\and challenge
\and tournament
\and AGI
\and AI
\and deep learning
\and NLP
\and Generative AI
\and analysis
\and game
\and grid-based
\and text-based
\and strategic
\and Tic-Tac-Toe
\and Connect Four
\and Gomoku
\and decision-making
\and prompt engineering
\and list
\and illustration
\and image
\and Anthropic
\and Claude
\and Gemini
\and GPT4
\and GPT4-o
\and Gemini-Pro
\and Gemini-Flash
\and Meta
\and LlaMA
}

\section{Introduction}
Recent advancements in large language models (LLMs) have marked significant progress in the field of Artificial Intelligence (AI) \cite{OverviewOfLLMs}. These developments prompt questions about the potential for achieving Artificial General Intelligence (AGI) \cite{AGI-Definition} and the timeline for such advancements. Predictions on the timeline for AGI vary \cite{NVIDIA-AGI}, \cite{Meta-AGI}, with some experts suggesting its inevitability \cite{Ilya-AGI}. A critical challenge in the journey towards AGI is developing benchmarks to assess AI’s evolving intelligence.

In this study, we introduce a novel and extensible benchmark for LLMs using grid-based games such as Tic-Tac-Toe, Connect Four, and Gomoku, utilizing three distinct types of prompts (list, illustration, image). This benchmark helps assess the capabilities of LLMs, including rule comprehension, strategic thinking, and the ability to process and understand complex text and image prompts. The benchmark provides open-source code for simulating these board games among LLMs and generating data files that store details of the simulated games. This study also includes the analysis of a total of 2,310 games played among leading LLMs, including Claude 3.5 Sonnet and Claude 3 Sonnet by Anthropic, Gemini 1.5 Pro and Gemini 1.5 Flash by Google, GPT-4 Turbo and GPT-4o by OpenAI, and Llama3-70B by Meta. The open-source game simulation code can be utilized to test other LLMs and prepare submission data for the leaderboard. The benchmark and leaderboard are designed to be extensible, accommodating new games and data submissions. The authors encourage contributions and welcome new results from other LLMs.

\section{Background and Related Research}
The deep learning revolution has profoundly transformed natural language processing (NLP) since the 2010s, with the introduction of the Transformer architecture in 2017 \cite{Attention} playing a pivotal role in this evolution. The Transformer architecture enabled parallel word processing, significantly improving the efficiency and handling of long-range text dependencies. This innovation led to the creation of models like BERT (Bidirectional Encoder Representations from Transformers) \cite{BERT} and OpenAI’s GPT (Generative Pre-trained Transformer) series \cite{GPT}. BERT advanced context under-standing by analyzing word relationships within sentences, while the GPT series excelled in generative language capabilities \cite{OverviewOfLLMs}.

The scale of LLMs expanded exponentially, resulting in models with billions of parameters and exceptional performance across various NLP tasks. Recent models include GPT-4 by OpenAI \cite{GPT4}, Gemini by Google \cite{GEMINI}, Claude by Anthropic \cite{CLAUDE}, Grok by xAI \cite{GROK}, and open-source options like LLaMA by Meta \cite{META} and Mistral by Mistral \cite{MISTRAL}. These models have pushed the boundaries of what is possible with LLMs, showcasing significant advancements in the field. LLMs are employed in diverse tasks such as text summarization, language translation, content generation, and question-answering \cite{APP-LLM}.

\subsection{Large Language Model Benchmarks}

LLMs produce outputs such that their responses can vary even with identical input \cite{LLM-Output}. Traditional metrics like accuracy, precision, F1 score, and mean squared error (MSE) are not suitable for evaluating LLM performance. Instead, specialized datasets and benchmarks are needed to assess LLM capabilities comprehensively \cite{LLM-Evaluation}.

Benchmarks such as GLUE \cite{GLUE}, SuperGLUE \cite{SUPERGLUE}, HELM \cite{HELM}, MMLU \cite{MMLU}, BIG-bench \cite{BIGBENCH}, ARC \cite{ARC}, TruthfulQA \cite{TruthfulQA}, HellaSwag \cite{HellaSwag}, and LiveBench \cite{LiveBench} provide diverse tasks that test various aspects of LLMs. GLUE, introduced in 2018, includes tasks like sentiment analysis and question-answering to evaluate natural language understanding. SuperGLUE, launched in 2019, extends GLUE with more demanding tasks such as multi-sentence reasoning and complex reading comprehension. The Massive Multitask Language Understanding (MMLU) benchmark tests LLMs across a wide array of subjects, including mathematics, history, computer science, and law, requiring extensive world knowledge and problem-solving abilities \cite{MMLU}. BIG-bench offers 204 varied tasks in areas such as linguistics, mathematics, reasoning, biology, and software development, allowing researchers to evaluate LLMs comprehensively while managing operational costs \cite{BIGBENCH}. HELM emphasizes transparency and performance in specific tasks, using a multi-metric approach that includes fairness, bias, and toxicity assessments. It continually adapts to add new scenarios, metrics, and models \cite{HELM}. The AI2 Reasoning Challenge (ARC) from the Allen Institute for Artificial Intelligence assesses AI systems' complex reasoning capabilities through multiple-choice questions. The ARC includes an Easy Set for basic retrieval methods and a Challenge Set for advanced reasoning, pushing AI towards deeper knowledge-based understanding \cite{ARC}. The TruthfulQA benchmark assesses the accuracy and truthfulness of LLM responses, specifically designed to measure how well models can generate accurate answers and avoid hallucinations \cite{TruthfulQA}. The HellaSwag benchmark tests common sense reasoning by presenting models with sentences and multiple possible endings, requiring them to choose the most logical continuation \cite{HellaSwag}. LiveBench addresses the test set contamination issue in LLM evaluation by offering a benchmark immune to such contamination and biases from human or LLM judging \cite{LiveBench}. It features frequently updated questions from recent sources, automatic scoring against objective ground-truth values, and diverse tasks spanning math, coding, reasoning, language, instruction following, and data analysis \cite{LiveBench}. 

The recent survey papers have provided comprehensive frameworks and methodologies that are invaluable for developing robust benchmarks. One such paper, "A Survey on Evaluation of Large Language Models," presents a detailed review focusing on three key dimensions: what to evaluate, where to evaluate, and how to evaluate \cite{SurveyEvaluationLLM}. It encompasses a broad spectrum of tasks, including general natural language processing, reasoning, medical applications, ethics, education, and more. This survey emphasizes the critical role of evaluation methods and benchmarks in assessing LLM performance, summarizing both successes and failures across different tasks, and highlighting future challenges in the field \cite{SurveyEvaluationLLM}. Similarly, "Evaluating Large Language Models: A Comprehensive Survey" categorizes LLM evaluation into knowledge and capability, alignment, and safety \cite{EvaluatingLLM}. This survey underscores the importance of rigorous assessment and the development of comprehensive evaluation platforms to ensure the safe and beneficial development of LLMs. It aims to guide responsible LLM advancement, ensuring that their evolution maximizes societal benefits while minimizing potential risks \cite{EvaluatingLLM}. Both surveys stress the necessity of treating evaluation as an essential discipline to aid the development of more proficient and ethically sound LLMs \cite{AndrewNG-NEED}.

\subsection{Utilizing Games for Evaluating LLMs}
Existing benchmarks primarily focus on language understanding tasks such as sentiment analysis, question-answering, and comprehension. Although some tasks in BIG-bench involve game-like problem-solving skills, they do not assess LLMs' performance in conventional games like chess or Go, which are valuable for evaluating strategic thinking and decision-making abilities. Using games as a benchmarking tool provides a unique perspective on LLM capabilities, highlighting their proficiency in understanding rules, formulating strategies, and making decisions. Strategic games like chess and Go emphasize predicting opponents' moves, while games involving linguistic interaction test language mastery and contextual understanding. The dynamic nature of games allows researchers to observe LLMs' adaptability and learning in real-time.

Engaging in gameplay offers a standardized framework for comparing various LLMs' performances under the same conditions, evaluating their strategic and creative problem-solving abilities, and capacity for innovative solutions. The controlled environment of games is instrumental for safely testing LLMs, allowing researchers to observe behaviors and mitigate potential risks or ethical concerns. Games involving human–AI interaction reveal how LLMs collaborate with or compete against humans, shedding light on human–AI relationship dynamics. Therefore, testing LLMs within the gaming domain extends beyond evaluating their ability to play games; it offers a comprehensive examination of strategic thinking, language processing, creativity, and adaptability, which is crucial for advancing AI research and ensuring the responsible development and deployment of these technologies.

Text-based games present a distinctive and challenging domain for benchmarking LLMs. These interactive fiction games require models to understand natural language, interpret evolving game states, and generate appropriate commands within narrative-driven environments, demanding a profound grasp of language, context, and strategic application \cite{GPT4}. Studies on models like FLAN-T5, Turing, and OPT in the text-based game "Detective" reveal that these LLMs fall short of state-of-the-art or human performance levels, facing challenges in adapting to game dynamics, learning from past interactions, and goal-oriented processing \cite{Text-BasedGames}.

The "GameEval-Evaluating LLMs on Conversational Games" paper introduces a framework for assessing LLMs through goal-driven conversational games, highlighting their abilities in complex discussions, decision-making, and problem-solving \cite{GameEval}. The SmartPlay benchmark assesses LLMs across diverse games, emphasizing their evolution as intelligent agents \cite{SmartPlay}. The MindAgent infrastructure evaluates multi-agent collaboration, enhancing human–AI coordination \cite{MidAgent}.

Studies on LLM behavior in social interaction games like the iterated Prisoner's Dilemma and the Battle of the Sexes show challenges in adapting to strategies requiring mutual understanding and flexibility \cite{PlayingGames}. Research by Lorè and Heydari on "Strategic Behavior of Large Language Models" underscores the role of context in strategic decision-making \cite{StrategicLLM}. Tsai et al. highlight limitations in LLMs like ChatGPT and GPT-4 in constructing world models and leveraging knowledge in text-based games, suggesting the potential for targeted benchmarks \cite{TextBasedLLM}.

The study "Can Large Language Models Serve as Rational Players in Game Theory?" evaluates LLMs' potential in game theory, identifying gaps in mimicking human rationality \cite{RationalPlayer}. Another study explores models like Claude 2, GPT-3.5, and GPT-4 in processing game strategy and spatial information through Tic-Tac-Toe, finding that prompt design significantly impacts performance \cite{SpatialLLM}.

GTBENCH evaluates LLMs' strategic reasoning in competitive game-theoretic tasks \cite{GTBENCH}. It features 10 tasks covering complete vs. incomplete information, dynamic vs. static, and probabilistic vs. deterministic scenarios. Results show LLMs struggle in complete, deterministic games but perform better in probabilistic ones. Commercial LLMs like GPT-4 outperform open-source models such as CodeLlama-34b-Instruct. Code-pretraining aids strategic reasoning, but advanced methods like Chain-of-Thought (CoT) and Tree-of-Thought do not consistently help. Detailed error profiles are provided to understand LLM behaviors \cite{GTBENCH}.

GAMEBENCH evaluates strategic reasoning in LLMs across nine game environments, each highlighting key reasoning skills \cite{GAMEBENCH}. Using GPT-3 and GPT-4, along with Chain-of-Thought prompting and Reasoning Via Planning (RAP), the study finds that while these frameworks improve performance, no model matches human capabilities, with GPT-4 sometimes performing worse than random actions. The games are selected to avoid overlap with the models' pretraining corpuses.

In a previous study, the authors evaluated the strategic thinking capabilities of various LLMs, including Claude 2.1, Gemini-Pro 1.0, GPT-3.5-Turbo, GPT-4, Llama2-70B, and Mistral Large, by having them play Tic-Tac-Toe through a mobile app \cite{Topsakal-Benchmark-TicTacToe}. This study builds upon that research with additional games, more in-depth analysis, and a user-friendly web-based game simulation software to evaluate more recent LLMs.

A recent survey paper explores the state of the art in applying LLMs to gaming, identifying the various roles LLMs can play within games. It highlights underexplored areas and promising directions for future research, reconciling the potential and limitations of LLMs in the gaming domain. This survey aims to serve as a foundation for future research and innovation in this emerging field \cite{LLMGameSurvey}.

Another recent survey paper explores LLM-based game agents and their role in advancing toward AGI. It introduces the conceptual architecture centered on perception, memory, thinking, role-playing, action, and learning. The paper reviews methodologies and adaptation agility of existing LLM-based game agents across six game genres: adventure, communication, competition, cooperation, simulation, and crafting \& exploration. It also provides future research directions in this field \cite{SurveyLLMGameAgent}.

These studies collectively deepen our understanding of LLMs' strengths and weaknesses in gaming and interactive contexts, providing a foundation for future research to enhance their performance and cognitive skills. They highlight the value of using games as benchmarks to expose the capabilities and limitations of current AI systems, paving the way for developing advanced models with sophisticated reasoning and strategic thinking.

\section{Methodology}
We have developed a benchmark to evaluate the capabilities of LLMs in rule comprehension and decision-making through grid-based games. This benchmark includes open-source web-based software for simulating games, accessible on GitHub \cite{GitHub}. The web application is built using JavaScript, HTML, and CSS, with server-side AWS Lambda functions written in Python to leverage LLMs hosted on AWS Bedrock. The game simulation web app enables LLMs to compete against each other, recording the details of each move for further analysis in JSON, CSV, TXT, and PNG formats, as well as summarizing game results.

\begin{figure}[htb]
  \centering
 \includegraphics[width=\textwidth,height=\textheight,keepaspectratio]{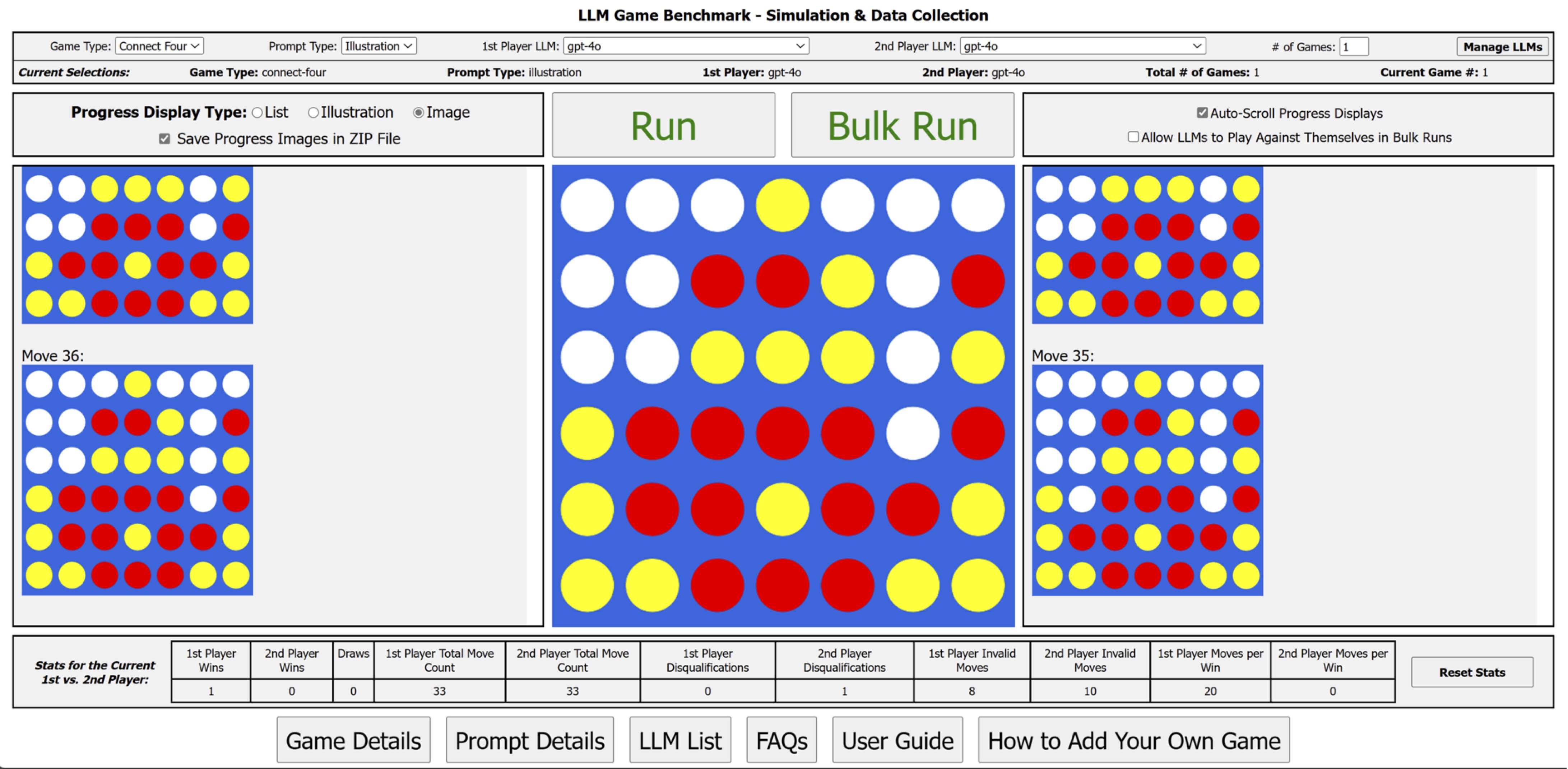}
  \caption{Web-based app for game simulation shows the progress of a Connect Four game.}
  \label{fig:GameSimulationUI}
\end{figure}

Currently, the benchmark includes Tic-Tac-Toe, Connect Four, and Gomoku, and is designed to be extensible to accommodate additional board games. A step-by-step guide for adding new game simulations is provided. As illustrated in Figure \ref{fig:GameSimulationUI}, the user interface of the game simulation web app allows users to select a game and the LLMs for the first and second players from a curated list. Users can also choose the type of predefined prompts (e.g., list, illustration, image) and specify the number of consecutive games for the selected game, prompt, and player combination.

\begin{figure}[htb]
  \centering
  \includegraphics[width=\textwidth,height=\textheight,keepaspectratio]{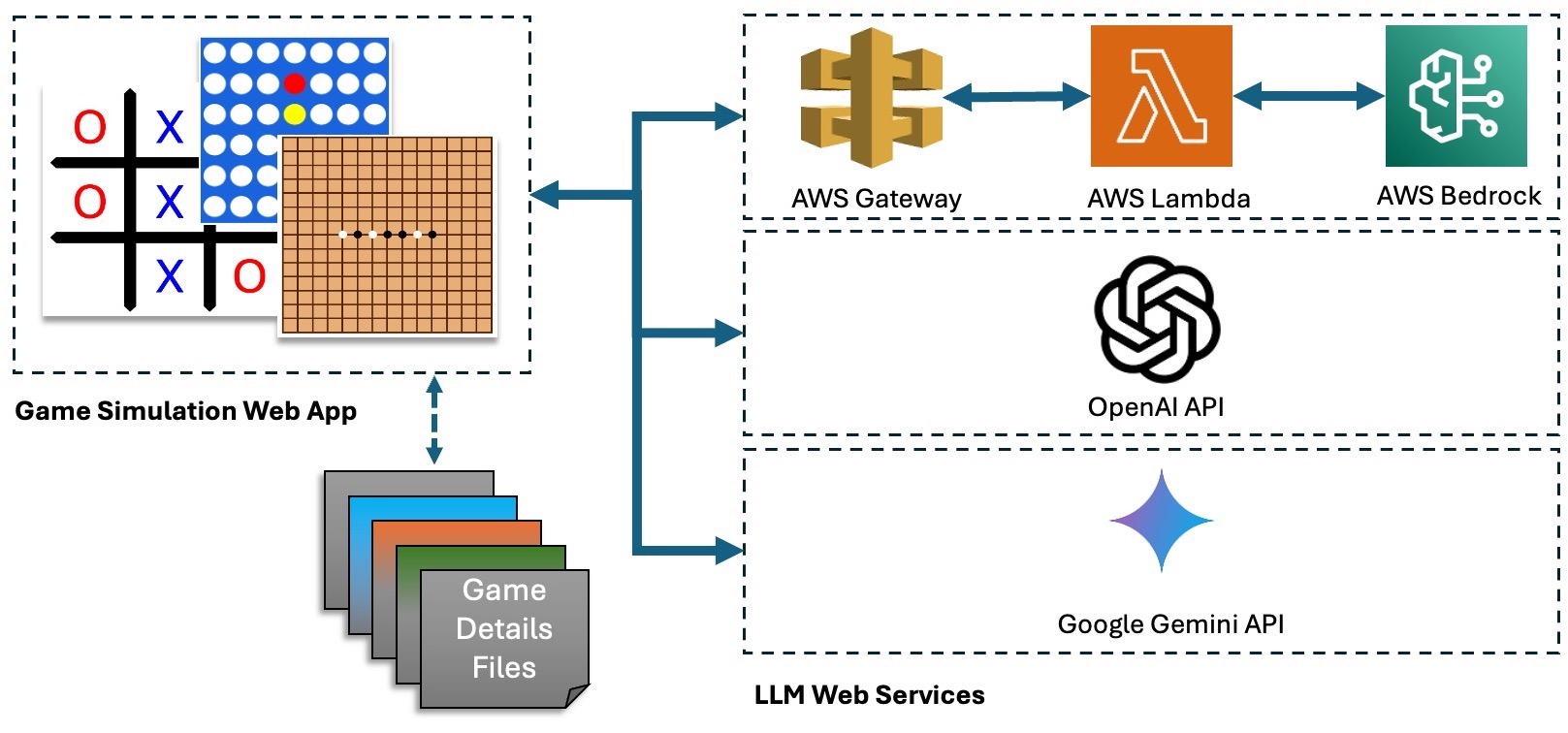}
  \caption{The illustration of web-based app and web service interactions to play a game.}
  \label{fig:WebServiceInteraction}
\end{figure}

The game simulation initiates by sending the selected prompt to the web API of the chosen LLM for the first player, then awaits its move. Upon receiving a response, the application updates the user interface to reflect the game's progress, as demonstrated in Figure \ref{fig:GameSimulationUI}, subsequently queries the chosen LLM for the second player, and awaits its move. The prompts, which include the current state of the game, are continuously sent to each LLM’s web service until a player wins, the game ends in a draw, or a player is disqualified for making invalid moves. Each query and response are recorded for every move. This methodology ensures seamless interaction between the application and the LLMs via web API calls. The interactions are illustrated in Figure \ref{fig:WebServiceInteraction}.

\subsection{Games Available on the Benchmark and Possibility of Expansion to New Games}

We have utilized three games in the benchmark; Tic-Tac-Toe, Connect Four and Gomoku. All of these games are classical two-player games played on a grid; 3 by 3, 6 by 7, and 15 by 15, respectively \cite{TicTacToe} \cite{Connect4} \cite{Gomoku}. These games can be adapted to larger grids. The explanations of these games are given in Table \ref{tbl:game-explanation} and the same explanations are used in the prompts.

Tic-Tac-Toe, Connect Four, and Gomoku are all solved games meaning their outcome (win, lose, or draw) can be correctly predicted from any position, assuming that both players play perfectly. In Tic-Tac-Toe, optimal play from both participants guarantees a draw. The first player can always win with optimal play in Connect Four. In the Gomoku game, the first player is guaranteed to win with optimal play \cite{Gomoku-Solved}.

We designed this benchmark to be extensible, allowing for the addition of new games such as checkers and chess. The code is modular, facilitating the easy integration of additional games. Additionally, we prepared a step-by-step guide on how to add a new game to the benchmark, which can be found on the game simulation page under the ‘How to Add Your Own Game’ link. We encourage interested individuals to contribute to the development of the benchmark

\subsection{LLMs Tested \& New Result Submission to the Leaderboard}

Numerous LLMs are available for evaluation. To ensure a meaningful and comprehensive assessment, we carefully selected LLMs based on several criteria. Firstly, we chose LLMs that are not specifically trained for the games used in the benchmark. Although the training data of proprietary LLMs is not publicly disclosed, we assume they are not trained explicitly for any of the benchmark games. We prioritized well-known, high-performing LLMs developed by industry leaders such as OpenAI and Google, given their significant contributions to AI advancements. Additionally, we included LLMs from emerging startup companies that have gained attention in the AI community, such as Anthropic. To further enrich our evaluation, we aimed to test open-source models and included Meta’s Llama3-70B model in our evaluation. This selection covers a broad spectrum of innovative approaches, technological capabilities, and accessibility options.

The landscape of LLMs changes rapidly, with new models frequently emerging with improved capabilities. Therefore, we provide game simulation software in the benchmark that generates submission files and encourage new submissions to the leaderboard. Contributors can evaluate other LLMs by integrating their LLM web service URL or API keys, generating new results, and submitting them to the leaderboard. We believe the leaderboard will allow people to see the progress of LLMs in different games as the leaderboard continues to be updated.

Currently, the benchmark includes results and detailed files for the following LLMs: Claude 3.5 Sonnet and Claude 3 Sonnet from Anthropic, Gemini 1.5 Flash and Gemini 1.5 Pro from Google, GPT-4 Turbo and GPT-4o from OpenAI, and Llama3-70B from Meta. To access these models, we utilized the web APIs provided by Google and OpenAI for the Gemini and GPT-4 models. For accessing models from Meta and Anthropic, we employed Amazon Bedrock services, leveraging serverless AWS Lambda functions and API Gateways, as depicted in Figure \ref{fig:WebServiceInteraction}.

To evaluate the decision-making capabilities of LLMs compared to random play, we included an option to select ‘random play‘ as the opponent. This option generates random responses for each move. By testing all the LLMs against random play, we aim to determine the extent to which LLMs outperform random decision-making in game scenarios.

\subsection{Details of the Prompts}

We utilized three types of prompts: list, illustration, and image. Each prompt is divided into eight main components: 1) an explanation of the game, 2) an explanation of the format for the game status, 3) the current game status, 4) a definition of the LLM's role followed by a request for its next move, 5) an explanation of the response format, 6) an explanation of invalid moves, 7) a warning if the previous move was invalid, including an explanation of why it was deemed invalid, and 8) the current number of invalid moves made by the player, as well as the number of invalid moves until the player is dis-qualified. The current game status, the invalid move warning, and the invalid move counts are dynamically generated and updated as the game progresses. Table \ref{tbl:prompt-parts} presents the components of a 'list' type prompt for the Tic-Tac-Toe game. This standardized format ensures consistency in prompts throughout the game while allowing for dynamic updates of the game state.

\begin{table}[htb]
\rowcolors{2}{gray!10}{white}
\caption{Explanation of the games used in the prompts.}
\begin{tabular}{p{0.12\linewidth} | p{0.83\linewidth}}
\hline
\rowcolor{gray!10}
Tic-Tac-Toe & Tic-Tac-Toe is a two-player game played on a 3 by 3 grid. The first player uses X symbols, and the second player uses O symbols. Players take turns placing their symbols in an empty cell on the grid. The objective is to align three of your symbols either horizontally, vertically, or diagonally. The player who first aligns three of their symbols wins the game. Strategic placement is crucial; besides aiming to align their symbols, players must also block their
opponent\textquotesingle s potential alignments to avoid defeat. \\ \hline
Connect Four & Connect Four is a two-player game played on a 6 by 7 grid. The first player uses red (R) discs, and the second player uses yellow (Y) discs. Players take turns dropping their discs into a column from the top row where there is still at least one empty space. The dropped disc falls straight down, occupying the lowest available row within the column. The objective is to align four of your discs either horizontally, vertically, or diagonally. The player who first aligns four of their discs wins the game. Strategic placement is crucial; besides aiming to align their discs, players must also block their opponent\textquotesingle s potential alignments to avoid defeat.\\ \hline
Gomoku & Gomoku is a two-player game played on a 15 by 15 grid. The first player uses black (B) dots, and the second player uses white (W) dots. Players take turns placing their dots on an empty intersection of the grid. The objective is to align five of your dots either horizontally, vertically, or diagonally. The player who first aligns five of their dots wins the game. Strategic placement is crucial; besides aiming to align their dots, players must also block their
opponent\textquotesingle s potential alignments to avoid defeat.\\ \hline
\end{tabular}
\label{tbl:game-explanation}
\end{table}

\begin{table}[htb]
\rowcolors{2}{gray!10}{white}
\caption{The parts of a prompt for the TicTacToe game. This table shows sample parts for the `list' type of prompt. The differences in the `illustration' and `image' prompts are given in Table \ref{tbl:prompt-differences}.}
\begin{tabular}{p{0.29\linewidth} | p{0.67\linewidth}}
\rowcolor{gray!10}
\hline
\textbf{Part} & \textbf{Prompt Content} \\ \hline
The explanation of the game. & \emph{Same as the corresponding game explanation given in Table \ref{tbl:game-explanation}} \\ \hline
The explanation of the format for the status of the game. The same for every game for the selected prompt type. The sample on the right is for the `list' type of prompt. & The current state of the game is recorded
in a specific format: each occupied location is delineated by a semicolon (\textquotesingle;\textquotesingle), and for each occupied location, the row number is listed first, followed by the column number, separated by a comma (\textquotesingle,\textquotesingle). If no locations are occupied by a player,
\textquotesingle None\textquotesingle{} is noted. Both the row and column numbers start from 1, with the top left corner of the grid indicated by 1,1. \\ \hline
The current game status. Dynamically generated. The sample on the right shows the current state for the `list' type of prompt. & The current state of the game is as follows:

The locations occupied by the first player: 1,1; 1,2; 3,2.

The locations occupied by the second player: 2,2; 3,3. \\ \hline
Defining the role of the LLM and then asking its next move. The same for every game. & You are an adept strategic player, aiming to win the game
in the fewest moves possible. You are the first (second) player. What would be your next move? \\ \hline
The explanation of the response format. & Suggest your next move in the following JSON format: \{\textquotesingle row\textquotesingle:
RowNumber, \textquotesingle column\textquotesingle: ColumnNumber\}. Do not include any additional commentary in your response. Replace RowNumber and ColumnNumber with the appropriate numbers for your move. Both RowNumber and ColumnNumber start at 1 (top left corner is \{\textquotesingle row\textquotesingle: 1, \textquotesingle column\textquotesingle: 1\}). The maximum value for RowNumber and ColumnNumber is 3, as the grid is 3 by 3. \\ \hline
The explanation of the invalid moves. & Please note that your move will be considered invalid if your response does not follow the specified format, or if you provide a RowNumber or ColumnNumber that is out of the allowed range, or already occupied by a previous move. Making more than 3 invalid moves will result in disqualification. \\ \hline
The warning if the last move was invalid, including a copy of the previous move and an explanation of why the move was invalid. Dynamically generated. & Your previous response was
\textquotesingle\{"row": X, ``column": Y\}\textquotesingle. This move
was deemed invalid for the following reason: \textquotesingle Already
Taken\textquotesingle. Please adjust accordingly. \\ \hline
The current number of invalid moves, as well as the number of invalid moves left until disqualification. Dynamically generated. & You currently have X invalid move(s). Y more invalid moves will result in
disqualification. \\ \hline
\end{tabular}
\label{tbl:prompt-parts}
\end{table}

\begin{table}[htb]
\rowcolors{2}{gray!10}{white}
\caption{The differences between the list, illustration and image type of prompts for the TicTacToe game. The prompt content slightly changes for Connect Four and Gomoku games. Please refer to the GitHub page for sample prompts for Connect Four and Gomoku.}
\begin{tabular}{p{0.16\linewidth} | p{0.16\linewidth} | p{0.6\linewidth}}
\rowcolor{gray!10}
\hline
\textbf{Type} & \textbf{Part} & \textbf{Prompt Content} \\ \hline
\multirow{2}{*}{list} & The explanation of the format for the status of the game. The same for every game. & The current state of the game is recorded in a specific format: each occupied location is delineated by a semicolon (\textquotesingle;\textquotesingle), and for each occupied location, the row number is listed first, followed by the column number,
separated by a comma (\textquotesingle,\textquotesingle). If no
locations are occupied by a player,
\textquotesingle None\textquotesingle{} is noted. Both the row and column numbers start from 1, with the top left corner of the grid indicated by 1,1. \\ \cline{2-3}
& The current game status. Dynamically generated. & The current state of the game is as follows:

The locations occupied by the first player: 1,1; 1,2; 3,2.

The locations occupied by the second player: 2,2; 3,3. \\ \hline
\multirow{2}{*}{illustration} & The explanation of the format for the status of the game. The same for every game. & The current state of the game is illustrated on a 3 by 3 grid.
\textquotesingle X\textquotesingle{} represents positions taken by the first player and \textquotesingle O\textquotesingle{} represents positions taken by the second player, while
\textquotesingle e\textquotesingle{} indicates an empty (available) position. \\ \cline{2-3} 
& The current game status. Dynamically generated. & The current state of the game is as follows:

eXe

eeO

eOe \\ \hline
\multirow{2}{*}{image} & The explanation of the format for the status of the game. The same for every game. & The current state of the game is depicted in an image showing a 3 by 3 grid, where
\textquotesingle X\textquotesingle{} represents positions taken by the first player and \textquotesingle O\textquotesingle{} represents positions taken by the second player. \\ \cline{2-3} 
& The current game status. Dynamically generated. & The current state of the game is given in the attached image.

\emph{{[}Image is sent in base64 format{]}} \\ \hline
\end{tabular}
\label{tbl:prompt-differences}
\end{table}

The content of the three types of prompts is consistent, except for the representation of the current state of the game (previous moves). The ‘list’ prompt enumerates previous moves for each player in a “row, column” format. The ‘illustration’ prompt depicts the current state of the grid using specific symbols for the first and second players (X and O for Tic-Tac-Toe, R and Y for Connect Four, and B and W for Gomoku) and 'e' for empty cells. The ‘image’ prompt visualizes the current state by providing a snapshot of the game board. These differences are detailed in Table \ref{tbl:prompt-differences}.

LLMs use parameters like max tokens, temperature, top-p, and frequency penalty to fine-tune their outputs \cite{LLM-Survey} \cite{Survey-LLM}. Max tokens control length, temperature adjusts creativity, top-p limits word choices to balance creativity and coherence, and frequency penalty reduces repetition. These settings customize LLM responses for applications such as customer support and content creation. We use default configurations for all parameters except the prompt, trusting the creators' fine-tuning for optimal performance.

The games continued until one player won, a draw occurred, or a disqualification was necessary. To gather statistical data on the outcomes, each game between opponents was repeated five times. Disqualification occurred if a player made more than a specified number of invalid moves: three for Tic-Tac-Toe, six for Connect Four, and fifteen for Gomoku. A move was deemed invalid if the response did not follow the specified format, the provided RowNumber or ColumnNumber was out of range, or a move was made to an already occupied space. When a player made an invalid move, they were warned about the invalidity, provided with the reason (as shown in Table \ref{tbl:prompt-parts}), and asked to make their move again. Continuous invalid moves led to disqualification to ensure fairness and prevent indefinite delays.
During the game sessions conducted through the web application, data on gameplay was collected and stored in JSON, CSV, TXT, and PNG formats. Samples of these files are available on GitHub, along with zip files containing the complete data from bulk runs performed for the results presented here. The JSON files include comprehensive details such as date/time, players, game result, duration, and all moves, covering both valid and invalid attempts. They also include the current game status sent to the LLM and the responses received from the LLM for each move. Additionally, the JSON format is used for the leaderboard submissions. A streamlined summary of the game is available in a CSV file. The TXT file provides an illustrated representation of the game moves, while the PNG files display snapshots of the board after each move. All files generated during this study are publicly available on GitHub.

\subsection{Details of the Data Generated by the Game Simulation Web App}

The data generated by the game simulation web app is downloaded as a zip file either after a 'run' between two LLMs or after a 'bulk run' between all LLMs listed as first and second players. The zip file generated after a 'bulk run' includes all the files that would be produced for each LLM pair match, as well as single files with the '\_all' suffix that encompass the content of all corresponding files from each pair's match. Each generated file’s name is prefixed using the following format: \textit{'Game-Type\_PromptType\_FirstPlayerLLM\_SecondPlayerLLM\_Result\_DateTime'}. For example, \textit{'tic-tac-toe\_list\_gemini-1.5-pro\_gemini-1.5-flash\_winner1st\_240707-164940'}. The data is stored in JSON, CSV, TXT, and PNG formats. The CSV file includes the same data as the corresponding JSON file. The TXT file includes concise statistics of a game and an illustration of the game's progress in text format. If the 'Save Progress Images in ZIP File' box is checked on the game simulation page, PNG files showing snapshots of each move during the game will be generated as well. The JSON file with the '\_submission' suffix can be sent to the first author to add the results of the matches to the leaderboard page. Table 4 lists the data included in the JSON file that provides the details of a game and the JSON file that can be used to submit the results. The sample files and zip files generated during the 'bulk run' of data collection for the results presented in this study are available on the GitHub page \cite{GitHub}.

\begin{table}[!htb]
\rowcolors{2}{gray!10}{white}
\caption{The data included in the JSON files.}
\begin{tabular}{p{0.15\linewidth} | p{0.80\linewidth}}
\rowcolor{gray!10}
\hline
\textbf{JSON File} & \textbf{Content} \\ \hline
The main JSON file that includes detailed outcomes of the game. For
example, the name of the file could be
``tic-tac-toe\_list\_gemini-1.5-pro\_gemini-1.5-flash\_winner1st\_240707-164940.json''
& - UUID: Unique identifier for the game instance.

- DateTime: Timestamp indicating when the game was played.

- GameType: Type of game (tic-tac-toe, connect four, or gomoku).

- PromptType: Type of prompt used (list, illustration, or image).

- PromptVersion: Version of the prompt (date that the prompt was last modified).

- GameNumber: Sequential identifier for the game.

- Player1: LLM model name for the first player.

- Player2: LLM model name for the second player.

- Result: Outcome of the game (winner1st, winner2nd, draw,
disqualified1st, disqualified2nd).

- GameDuration: Duration of the game in seconds.

- TotalMoves: Total number of moves made during the game.

- Player1Moves: Number of moves by the first player.

- Player2Moves: Number of moves by the second player.

- Player1InvalidAlreadyTaken: Number of moves where the first player
attempted to place a move in an already occupied location.

- Player2InvalidAlreadyTaken: Number of moves where the second player attempted to place a move in an already occupied location.

- Player1InvalidFormat: Number of moves in invalid format by the first player.

- Player2InvalidFormat: Number of moves in invalid format by the second player.

- Player1OutOfBounds: Number of moves made outside the board boundaries by the first player.

- Player2OutOfBounds: Number of moves made outside the boundaries by the second player.

- FinalGameState: The final status of the board presented in the chosen prompt type.

- Moves: Array of move objects detailing each move made during the game. Each move object includes:

- - MoveNumber: Sequence number of the move.

- - Player: Indicates whether the move was made by Player 1 or Player 2.

- - Row: Row coordinate of the move on the grid-based board.

- - Column: Column coordinate of the move on the grid-based board.

- - Outcome: Result of the move (e.g., "Valid", "Already Taken").

- - CurrentStatus: The current status of the board sent to the LLM in the prompt format (list, illustration, or image). If the prompt type is image, it includes a base64-encoded string representing the game board\textquotesingle s state after the move.

- - Response: The response provided by the player, specifying the
move. \\ \hline
Submission (the file type has the suffix `\_submission.json') & - ProviderEmail: Email of the provider submitting
the results. This information can be entered in the
\textquotesingle Manage LLMs\textquotesingle{} settings on the game
simulation page.

- UUID: Unique identifier for the game instance.

- DateTime: Timestamp indicating when the game was played.

- GameType: Type of game (tic-tac-toe, connect four, or gomoku).

- PromptType: Type of prompt used (list, illustration, or image).

- PromptVersion: Version of the prompt (date that the prompt was last
modified).

- LLM1stPlayer: The LLM model name for the first player.

- LLM2ndPlayer: The LLM model name for the second player.

- WinRatio-1st: Win ratio of the first player.

- WinRatio-2nd: Win ratio of the second player.

- Wins-1st: Number of wins by the first player.

- Wins-2nd: Number of wins by the second player.

- Disqualifications-1st: Number of disqualifications for the first
player.

- Disqualifications-2nd: Number of disqualifications for the second
player.

- Draws: Number of games that ended in a draw.

- InvalidMovesRatio-1st: Ratio of invalid moves made by the first
player.

- InvalidMovesRatio-2nd: Ratio of invalid moves made by the second
player.

- TotalMoves-1st: Total number of moves made by the first player.

- TotalMoves-2nd: Total number of moves made by the second player. \\ \hline
\end{tabular}
\label{tbl:file-json}
\end{table}

\subsection{Metrics and Methods for Evaluation}

We evaluated the performance of LLMs across three games (Tic-Tac-Toe, Connect Four, and Gomoku) using different prompt types (list, illustration, and image) to assess their ability to handle various formats of game state representation. Performance comparisons were made against a random play strategy to establish a baseline, highlighting the strategic advantages of the LLMs. The primary metrics for evaluation included win rates, draw rates, and disqualification rates, providing an overview of the LLMs' performance as both the first and second players. Additionally, we tracked the number of invalid moves per game and the average number of moves per game to assess rule adherence and game engagement. To delve deeper into the LLMs' strategic thinking, we analyzed missed opportunities to win or block an opponent's win, counting instances where the LLMs failed to make critical moves. We presented the missed opportunities per game by averaging the missed opportunities across all games that resulted in a win, draw, or disqualification. We also normalized the number of missed opportunities by the number of valid moves to calculate the percentage of missed opportunities per valid move. The results were visualized through charts and tables to provide a clear depiction of performance metrics and trends, as shown in the Results section. Additionally, we present the outcomes of each match between seven LLMs and a random play generator across different games (a total of 2,310 matches) in a results matrix table. We also maintain a leaderboard on the GitHub page that allows for filtering and sorting results by different metrics. We encourage community contributions to suggest and implement new evaluation metrics and methodologies, fostering a collaborative approach to advancing the understanding of LLM capabilities.

\section{Results}
In this section, we present the outcomes of games played among LLMs. These results are based on data files generated by the open-source game simulation web software and shared on the GitHub page. 

\begin{figure}[htb]
  \centering
 \includegraphics[width=\textwidth,height=\textheight,keepaspectratio]{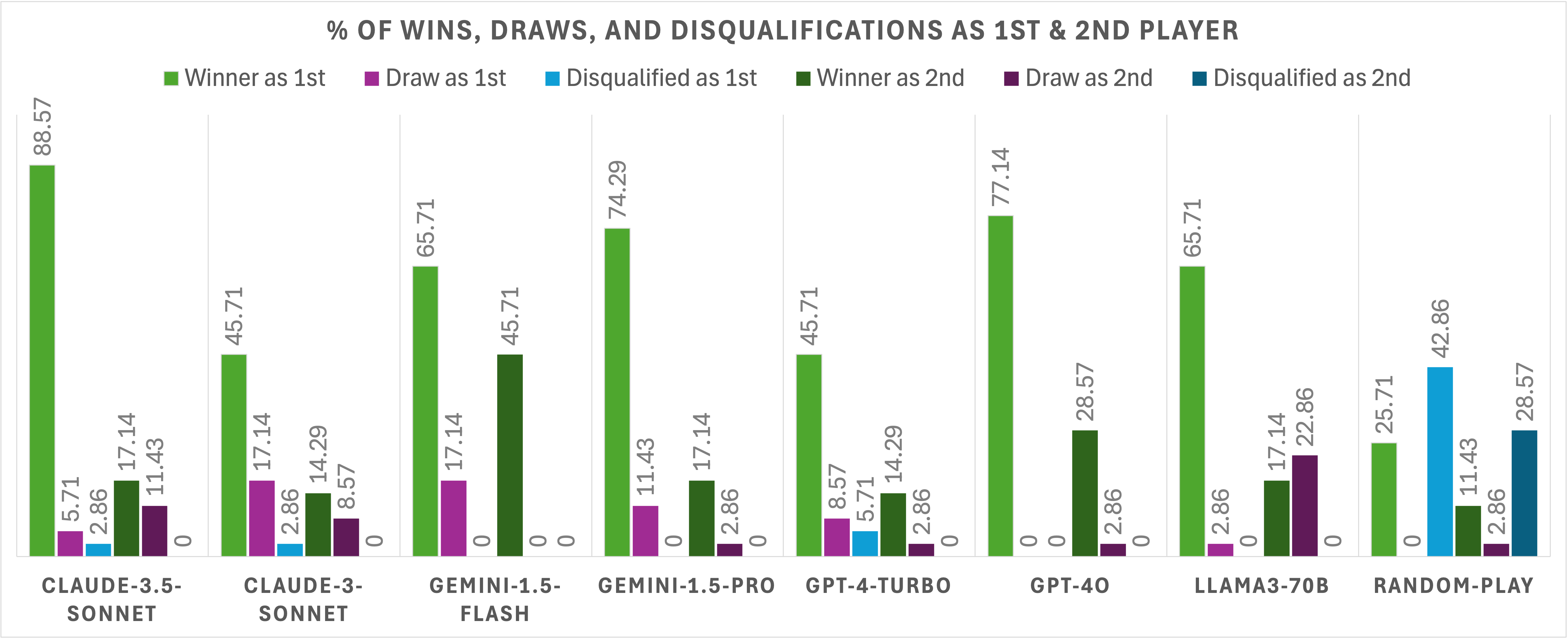}
  \caption{Tic-Tac-Toe game outcomes using the ‘list’ prompt where each LLM faced six others and the ‘random play’ as both player 1 and player 2, playing each opponent 5 times (280 games total).}
  \label{fig:TicTacToeList}
\end{figure}

Figure \ref{fig:TicTacToeList} displays the outcomes of Tic-Tac-Toe games using the list prompt type, where seven LLMs competed against others and a random play opponent, engaging in five matches per opponent for a total of 280 games. The chart summarizes the performance of the seven LLMs as well as random play in terms of win rates, draw rates, and dis-qualification rates as both the first and second players. Claude 3.5 Sonnet has the highest winning percentage as the first player (88.57\%) but a lower winning percentage as the second player (17.14\%). GPT-4o and Gemini 1.5 Pro show strong performance as both the first and second players, while random play results in the highest disqualification rates

\begin{figure}[htb]
  \centering
 \includegraphics[width=\textwidth,height=\textheight,keepaspectratio]{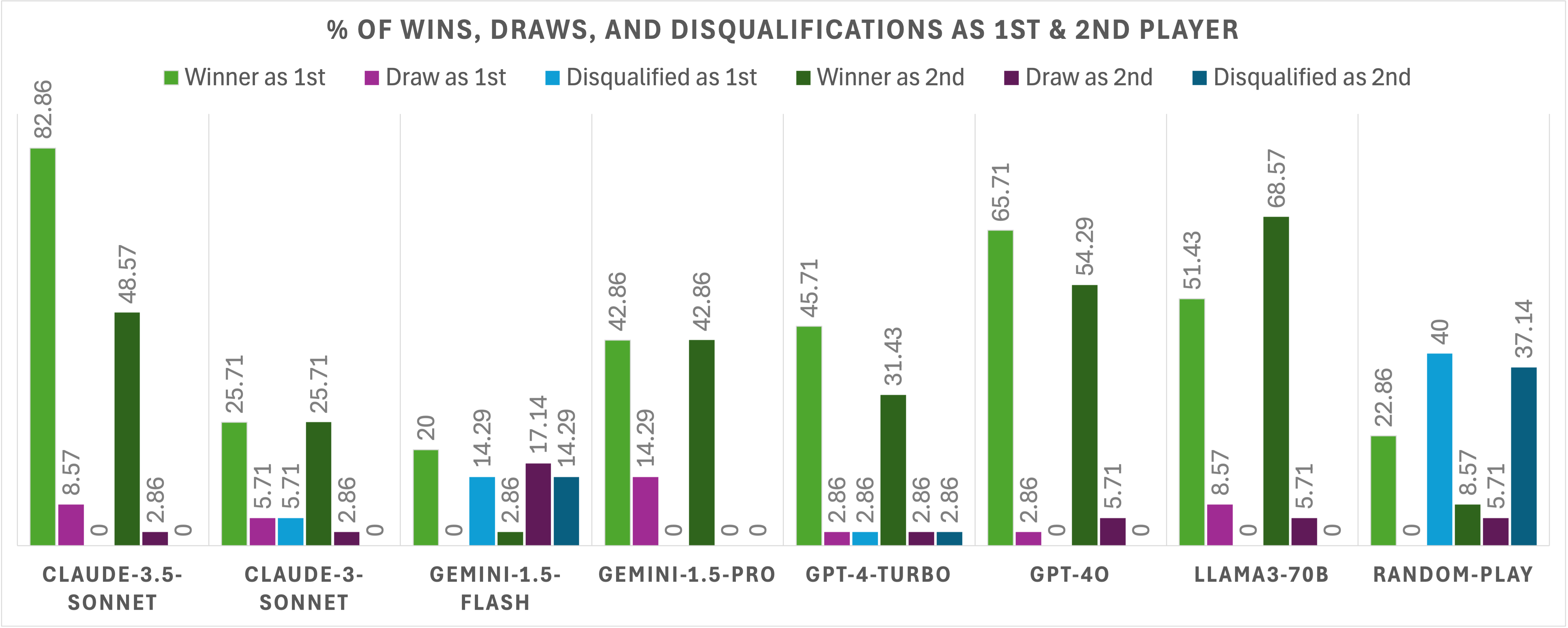}
  \caption{Tic-Tac-Toe game outcomes using the ‘illustration’ prompt where each LLM faced six others and the ‘random play’ as both player 1 and player 2, playing each opponent 5 times (280 games total).}
  \label{fig:TicTacToeIllustration}
\end{figure}

Figure \ref{fig:TicTacToeIllustration} displays the performance metrics of seven LLMs and a random play strategy in terms of win rates, draw rates, and disqualification rates when playing Tic-Tac-Toe as the first and second player, based on the illustration prompt format. The chart shows significant performance variations among the LLMs, with Claude 3.5 Sonnet exhibiting the highest winning rates as the first player, indicating a strong strategic advantage. Llama3-70B and GPT-4 Turbo also demonstrate strong performance. Disqualification rates are generally low but notable for some models, such as Gemini 1.5 Flash, indicating occasional invalid moves. The random player serves as a baseline comparison, with lower winning rates and higher disqualification rates, highlighting the superior strategic capabilities of the LLMs.

\begin{figure}[htb]
  \centering
 \includegraphics[width=\textwidth,height=\textheight,keepaspectratio]{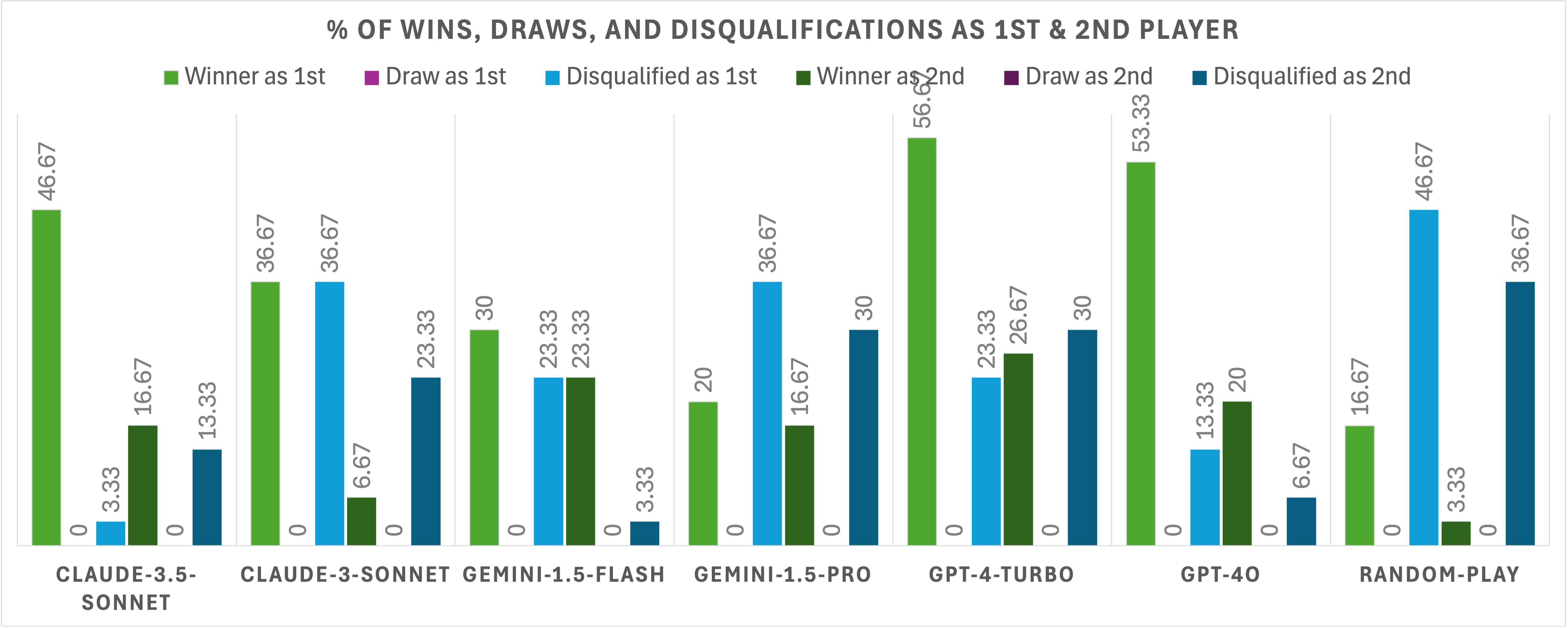}
  \caption{Tic-Tac-Toe game outcomes using the ‘image’ prompt where each LLM faced five others and the ‘random play’ as both player 1 and player 2, playing each opponent 5 times (210 games total).}
  \label{fig:TicTacToeImage}
\end{figure}

Figure \ref{fig:TicTacToeImage} displays the performance metrics of various LLMs and a random play strategy in terms of win rates, draw rates, and disqualification rates when playing Tic-Tac-Toe as the first and second player, based on the image prompt format. Key observations indicate that Claude 3.5 Sonnet has a high disqualification rate as both the first (46.67\%) and second player (53.33\%), indicating a struggle with rule compliance. Similarly, GPT-4 Turbo and Gemini 1.5 Pro also show significant disqualification rates. GPT-4 Turbo and GPT-4o exhibit the highest winning rates. The random play baseline has high disqualification rates, highlighting the strategic advantages of LLMs compared to random strategies. No draws occurred in the Tic-Tac-Toe games using the image prompt. Llama3-70B does not accept images, so it was not used when testing any of the games with the image prompt type.

\begin{figure}[htb]
  \centering
 \includegraphics[width=\textwidth,height=\textheight,keepaspectratio]{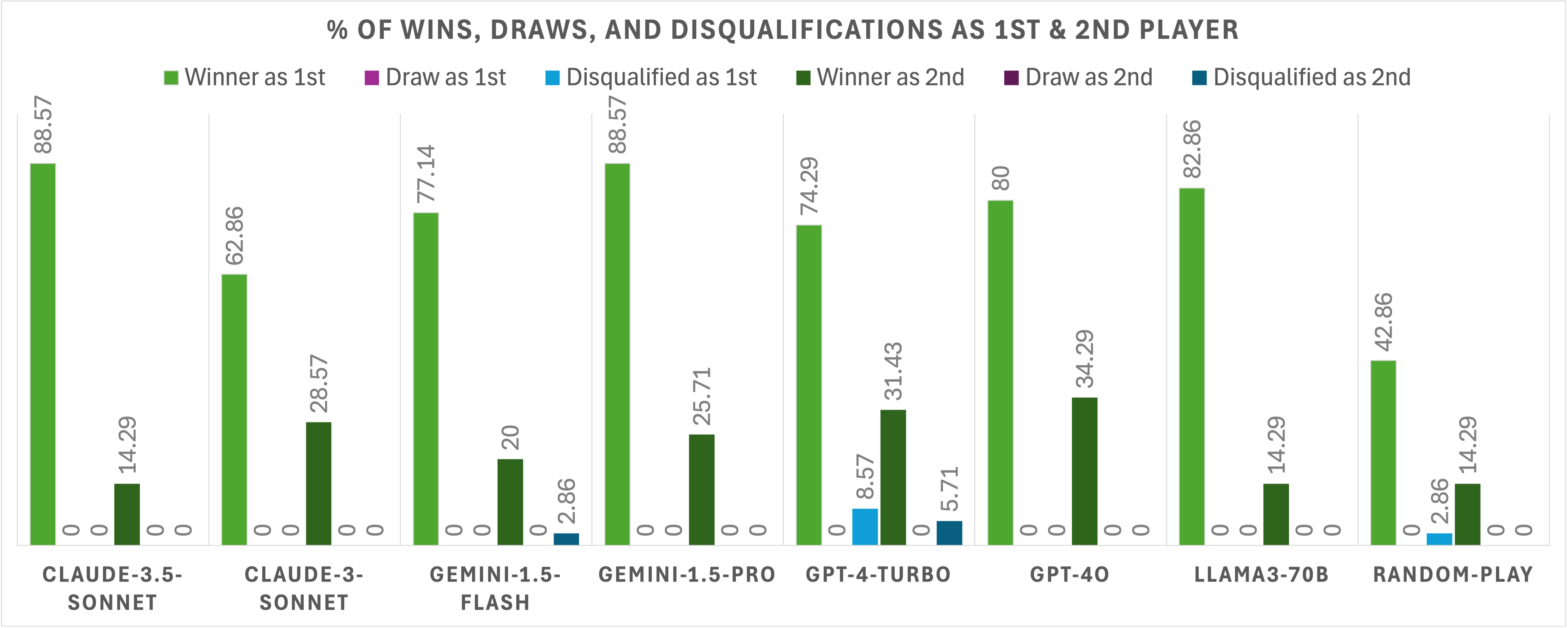}
  \caption{Connect Four game outcomes using the ‘list’ prompt where each LLM faced sic others and the ‘random play’ as both player 1 and player 2, playing each opponent 5 times (280 games total).}
  \label{fig:Connect4List}
\end{figure}

The chart in Figure \ref{fig:Connect4List} displays the performance metrics of various LLMs and a random play strategy in terms of win rates, draw rates, and disqualification rates when playing Connect Four as the first and second player using the list prompt type. Claude 3.5 Sonnet and Gemini 1.5 Pro show outstanding performance with a high winning rate of 88.57\% as the first player. Most LLMs demonstrated strong performance when considering their total win rates as both first and second players. The random player, serving as a baseline, has lower winning rates and some disqualifications, highlighting the strategic advantages of the LLMs. No draws occurred in the Connect Four games using the list prompt.

\begin{figure}[htb]
  \centering
 \includegraphics[width=\textwidth,height=\textheight,keepaspectratio]{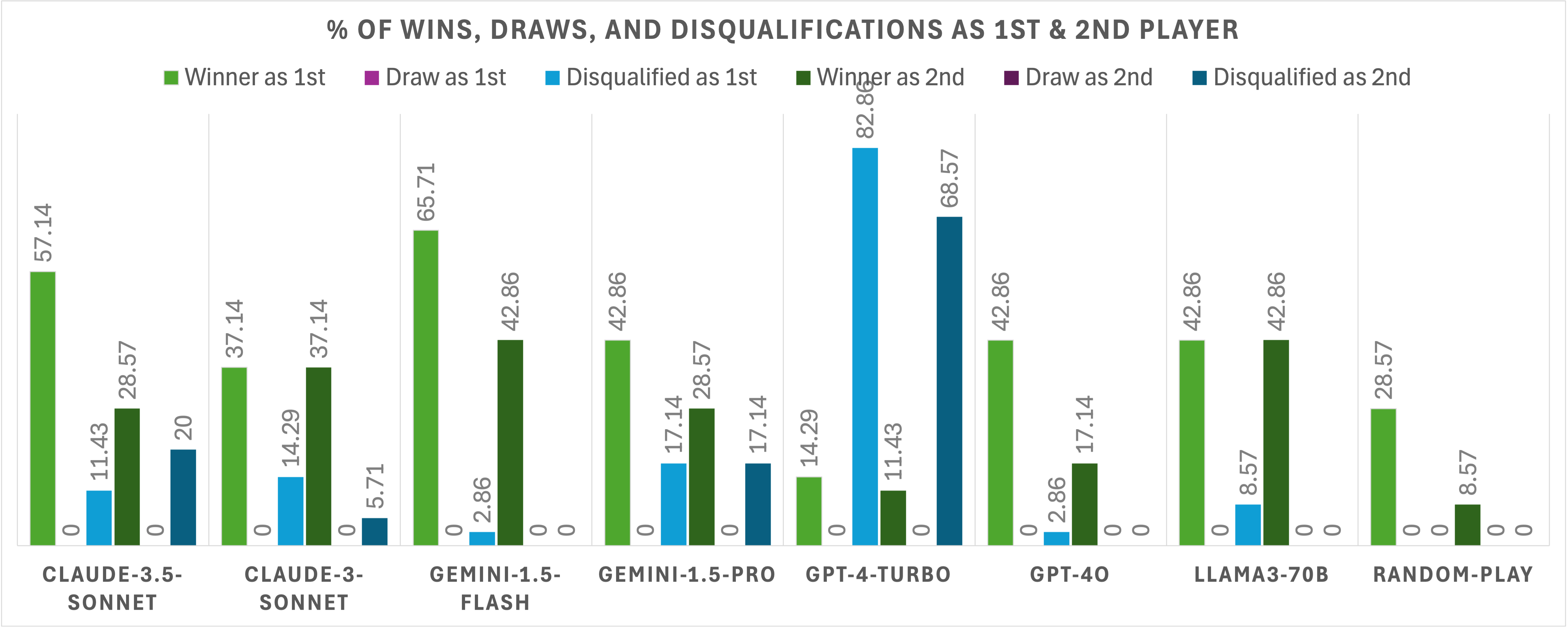}
  \caption{Connect Four game outcomes using the ‘illustration’ prompt where each LLM faced six others and the ‘random play’ as both player 1 and player 2, playing each opponent 5 times (280 games total).}
  \label{fig:Connect4Illustration}
\end{figure}

Figure \ref{fig:Connect4Illustration} presents the performance metrics of various LLMs and a random play strategy in terms of win rates and disqualification rates when playing Connect Four as the first and second player using the illustration prompt type. GPT-4 Turbo has the highest disqualification rates as both the first and second players. The random play, serving as a baseline, has the second lowest win rate as the first player and the lowest win rate as the second player. No draws occurred in the Connect Four games using the illustration prompt.

\begin{figure}[htb]
  \centering
 \includegraphics[width=\textwidth,height=\textheight,keepaspectratio]{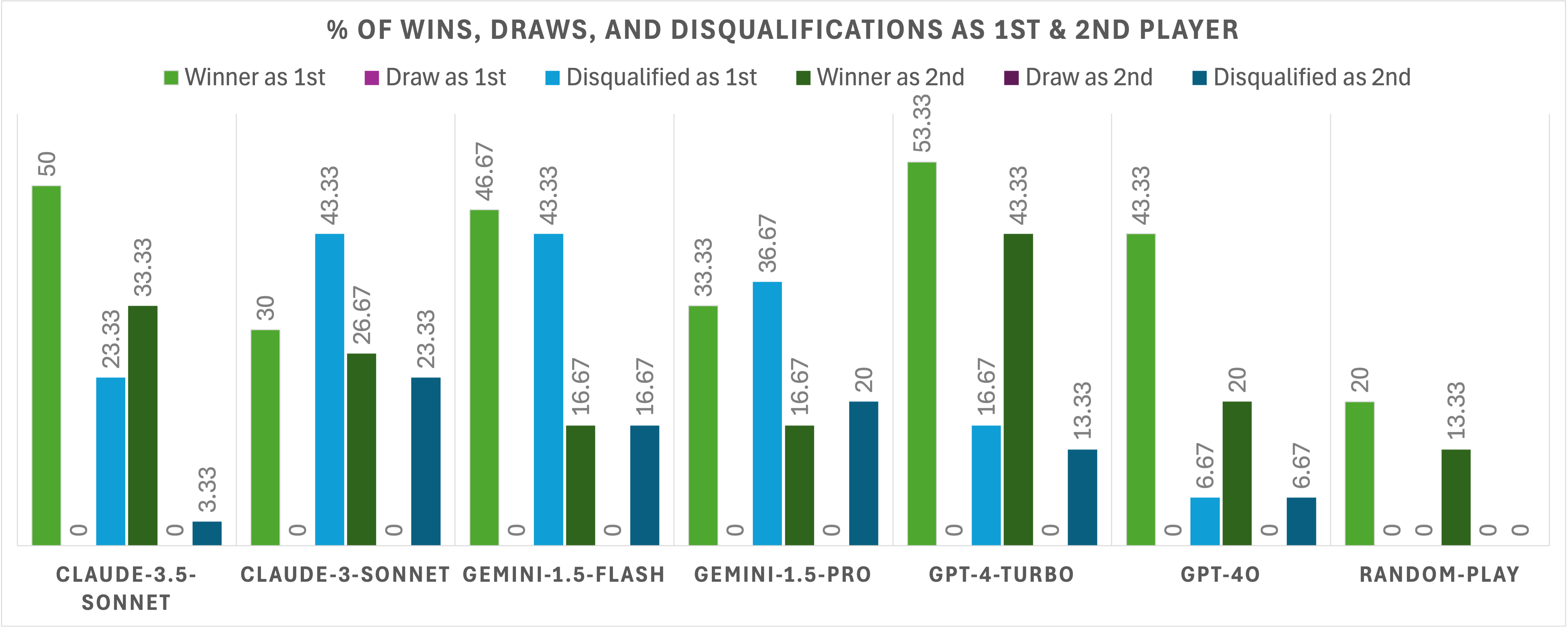}
  \caption{Connect Four game outcomes using the ‘image’ prompt where each LLM faced five others and the ‘random play’ as both player 1 and player 2, playing each opponent 5 times (210 games total).}
  \label{fig:Connect4Image}
\end{figure}

The chart in Figure \ref{fig:Connect4Image} illustrates the performance metrics of various LLMs and a random play strategy in terms of win rates and disqualification rates when playing Connect Four as the first and second player using the image prompt format. GPT-4 Turbo and Claude 3.5 Sonnet demonstrate strong winning performance as both the first and second players. Claude 3 Sonnet and Gemini 1.5 Flash have high disqualification rates overall. The random play baseline has the lowest winning rates. No draws occurred during the matches between the opponents.

\begin{figure}[htb]
  \centering
 \includegraphics[width=\textwidth,height=\textheight,keepaspectratio]{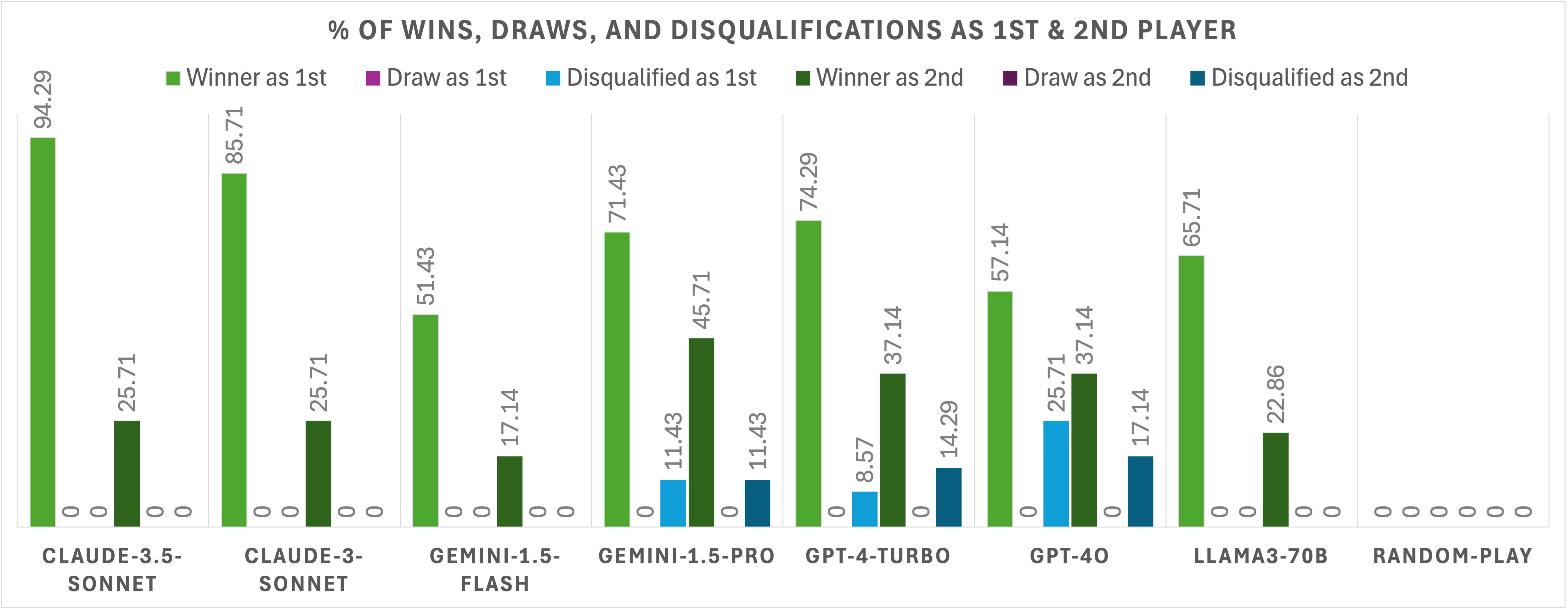}
  \caption{Gomoku game outcomes using the ‘list’ prompt where each LLM faced six others and the ‘random play’ as both player 1 and play-er 2, playing each opponent 5 times (280 games total).}
  \label{fig:GomokuList}
\end{figure}

Figure \ref{fig:GomokuList} displays the performance metrics of various LLMs and a random play strategy in terms of win rates and disqualification rates when playing Gomoku as the first and second player using the list prompt. Claude 3.5 Sonnet demonstrates exceptional performance with a 94.29\% win rate as the first player and 25.71\% win rate as the second player, with no disqualifications. Claude 3 Sonnet also performs well with an 85.71\% win rate as the first player and 25.71\% win rate as the second player, maintaining a clean record. Gemini 1.5 Pro has a high win rate of 71.43\% as the first player and 45.71\% as the second player but exhibits an 11.43\% disqualification rate as both the first and second player. GPT-4 Turbo stands out with a 74.29\% win rate as the first player and 37.14\% win rate as the second player, showing minimal disqualifications. GPT-4o performs well with a 57.14\% win rate as the first player and 37.14\% win rate as the second player, with some disqualifications as the first player. Llama3-70B exhibits a win rate of 65.71\% as the first player and 22.86\% as the second player, with no disqualifications. The random player, serving as a baseline, has no wins, draws, or disqualifications recorded.

\begin{figure}[htb]
  \centering
 \includegraphics[width=\textwidth,height=\textheight,keepaspectratio]{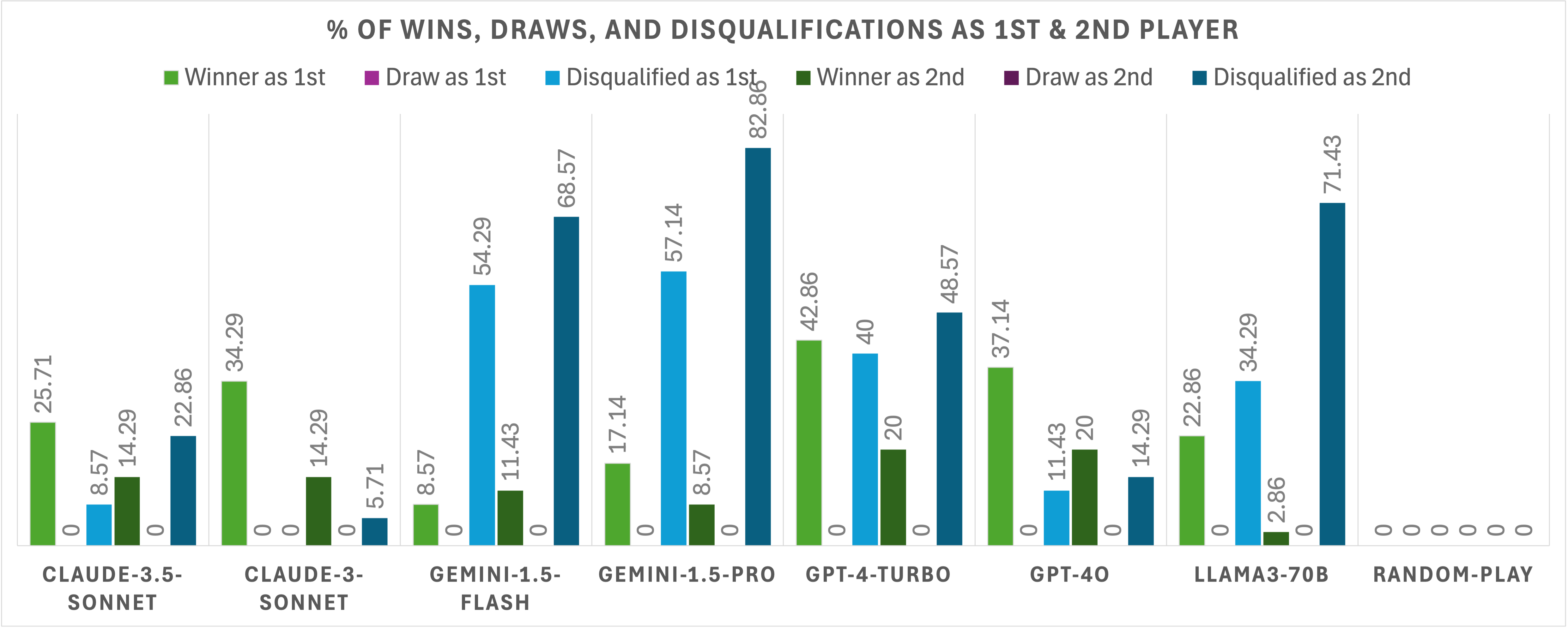}
  \caption{Gomoku game outcomes using the ‘illustration’ prompt where each LLM faced six others and the ‘random play’ as both player 1 and player 2, playing each opponent 5 times (280 games total).}
  \label{fig:GomokuIllustration}
\end{figure}

The chart in Figure \ref{fig:GomokuIllustration} displays the performance metrics of various LLMs and a random play strategy in terms of win rates and disqualification rates when playing Gomoku as the first and second player. Significant disqualification rates are evident among several LLMs, particularly when playing as the second player, indicating challenges in adhering to the game's rules. Models like Gemini 1.5 Flash and Llama3-70B had high disqualification rates. Winning rates vary widely, with some models showing strong performance as the first player while struggling as the second player. The notable disqualification rates suggest that strategic complexity and rule comprehension are significant factors affecting LLM performance. Overall, the chart highlights the variability in strategic abilities and reliability of different LLMs, emphasizing the need for further improvements in their rule adherence and strategic planning capabilities. The random play baseline, with no recorded wins, draws, or disqualifications, underscores the superior strategic thinking and performance of the LLMs despite their challenges.

\begin{figure}[htb]
  \centering
 \includegraphics[width=\textwidth,height=\textheight,keepaspectratio]{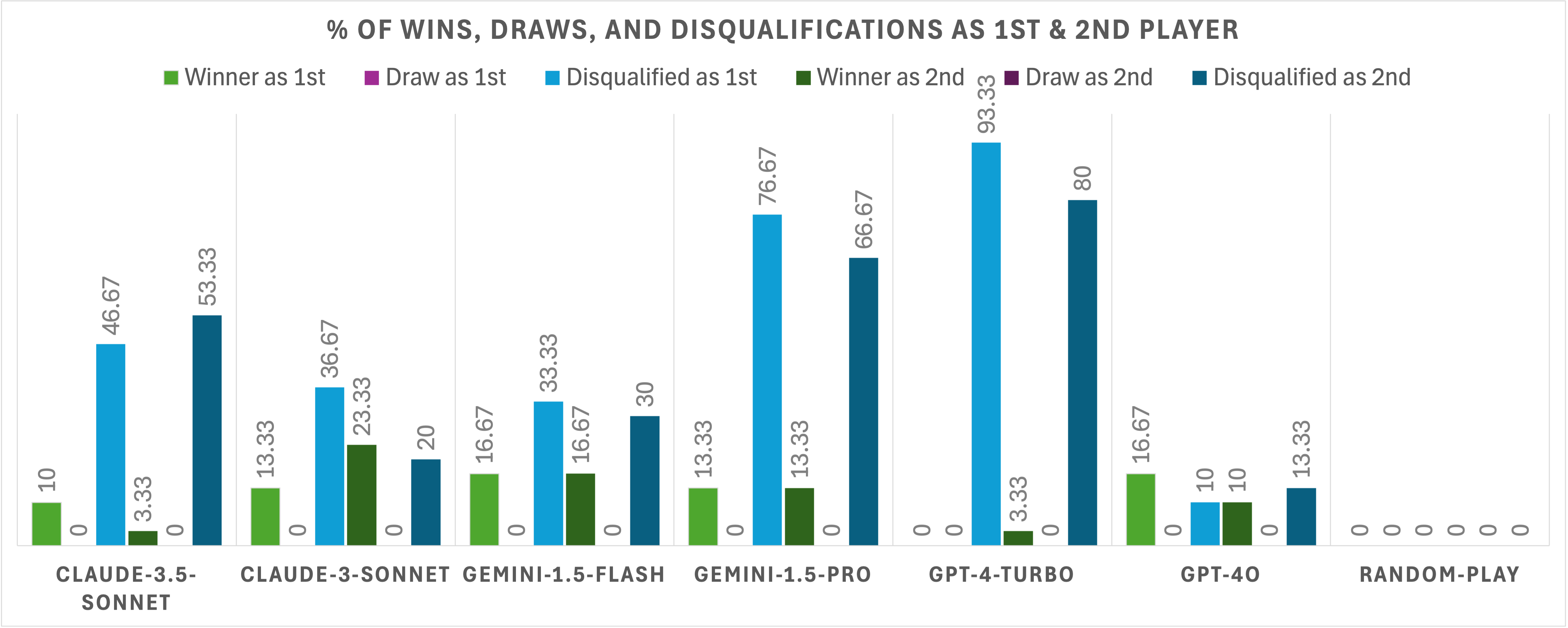}
  \caption{Gomoku game outcomes using the ‘image’ prompt where each LLM faced five others and the ‘random play’ as both player 1 and player 2, playing each opponent 5 times (210 games total).}
  \label{fig:GomokuImage}
\end{figure}

Figure \ref{fig:GomokuImage} illustrates the performance metrics of various LLMs and a random play strategy in terms of win rates and disqualification rates when playing Gomoku as the first and second player using the image prompt type. A notable pattern is the high disqualification rates for some models, particularly when playing as the second player, suggesting difficulties in rule adherence. There is a marked variation in win rates among the models, with some achieving higher success as the first player. The random player baseline demonstrates no recorded wins, draws, or disqualifications.

\begin{figure}[htb]
  \centering
 \includegraphics[width=\textwidth,height=\textheight,keepaspectratio]{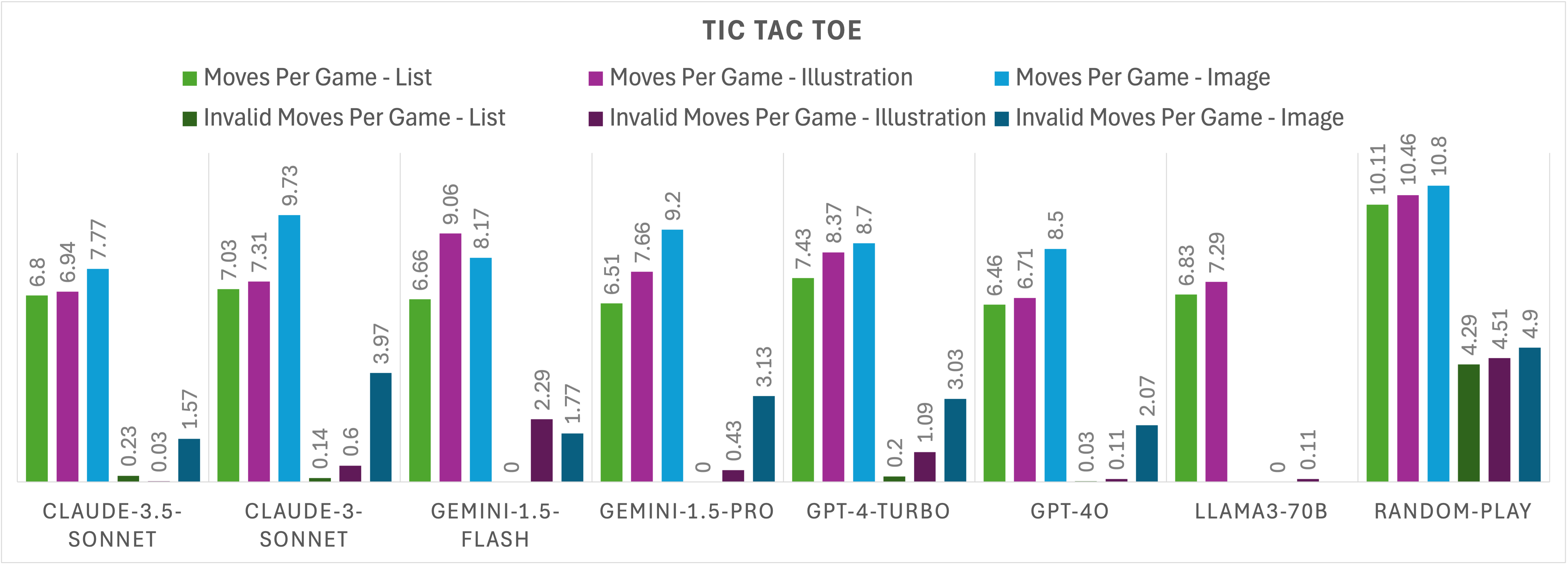}
  \caption{Moves per game and invalid moves (already taken) per game for Tic-Tac-Toe.}
  \label{fig:TicTacToeInvalidMoves}
\end{figure}

The chart in Figure \ref{fig:TicTacToeInvalidMoves} illustrates the performance of LLMs and a random play strategy in terms of moves per game and invalid moves per game when they participated as both first and second players in Tic-Tac-Toe across three prompt types: list, illustration, and image. The random play strategy serves as a baseline, indicating performance without strategic thinking. Generally, the number of moves per game increases with the complexity of the prompt, with image prompts resulting in the highest number of moves across all models. For list prompts, LLM performance ranged from 6.46 to 7.43 moves per game, while random play showed a higher number at 10.11. Illustration prompts saw a slight increase in moves for most models, peaking with Gemini 1.5 Flash. Invalid moves were minimal for list prompts but increased significantly for illustration prompts, particularly for Gemini 1.5 Flash, and were highest for image prompts, notably for Claude 3 Sonnet and Gemini 1.5 Pro. Random play consistently exhibited higher invalid moves across all prompt types, underscoring its lack of strategic planning compared to the LLMs. Llama3-70B was not used for the image prompt since it cannot accept images.

\begin{figure}[htb]
  \centering
 \includegraphics[width=\textwidth,height=\textheight,keepaspectratio]{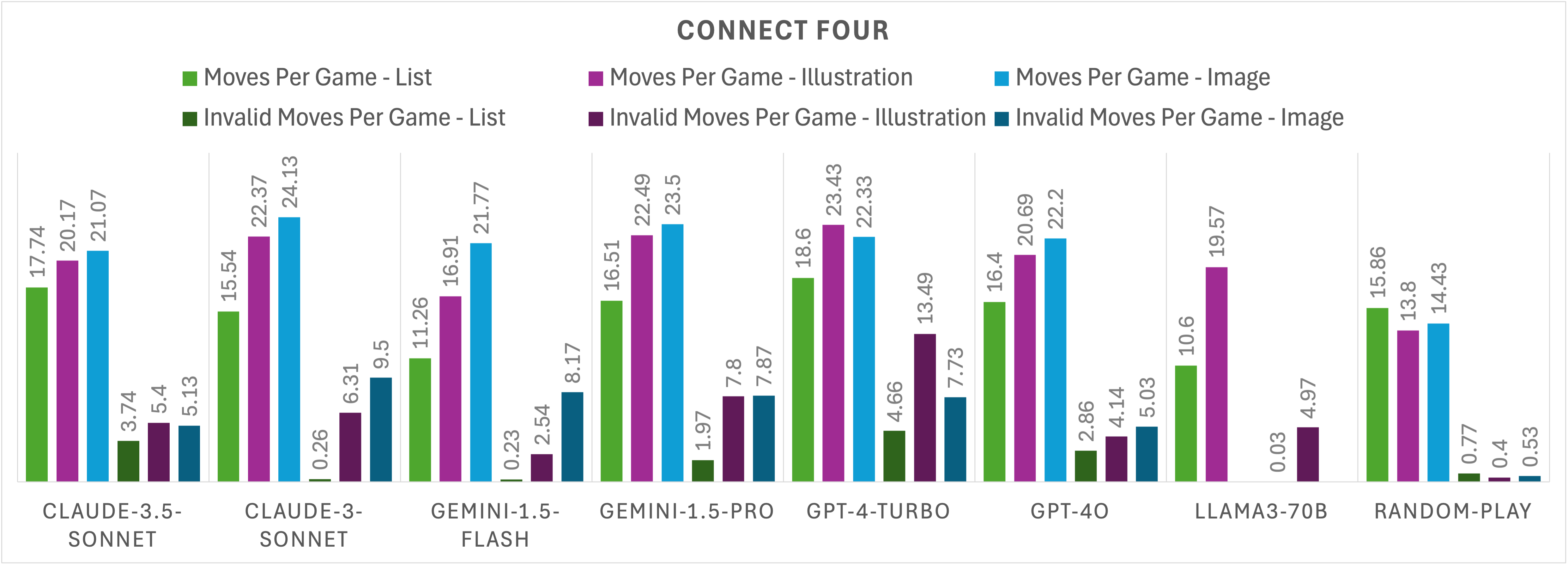}
  \caption{Moves per game and invalid moves (already taken) per game for Connect Four.}
  \label{fig:Connect4InvalidMoves}
\end{figure}

The chart in Figure \ref{fig:Connect4InvalidMoves} compares the performance of LLMs and a random play strategy in terms of moves per game and invalid moves per game, when they participated as both first and second players, for Connect Four across three prompt types: list, illustration, and image. Overall, the number of moves per game tends to increase with the complexity of the prompt, with the highest number of moves observed in the image prompt format. For list prompts, the LLMs demonstrated consistent moves per game with minimal invalid moves. However, there was a significant increase in invalid moves in the illustration prompts for GPT-4 Turbo and in the image prompts for Claude 3 Sonnet. Notably, random play showed relatively lower invalid moves, likely because invalid moves (already taken slots) can only occur when all rows of a column are filled in Connect Four. These results highlight the challenges faced by LLMs in handling more complex and visually demanding prompt formats. 

\begin{figure}[htb]
  \centering
 \includegraphics[width=\textwidth,height=\textheight,keepaspectratio]{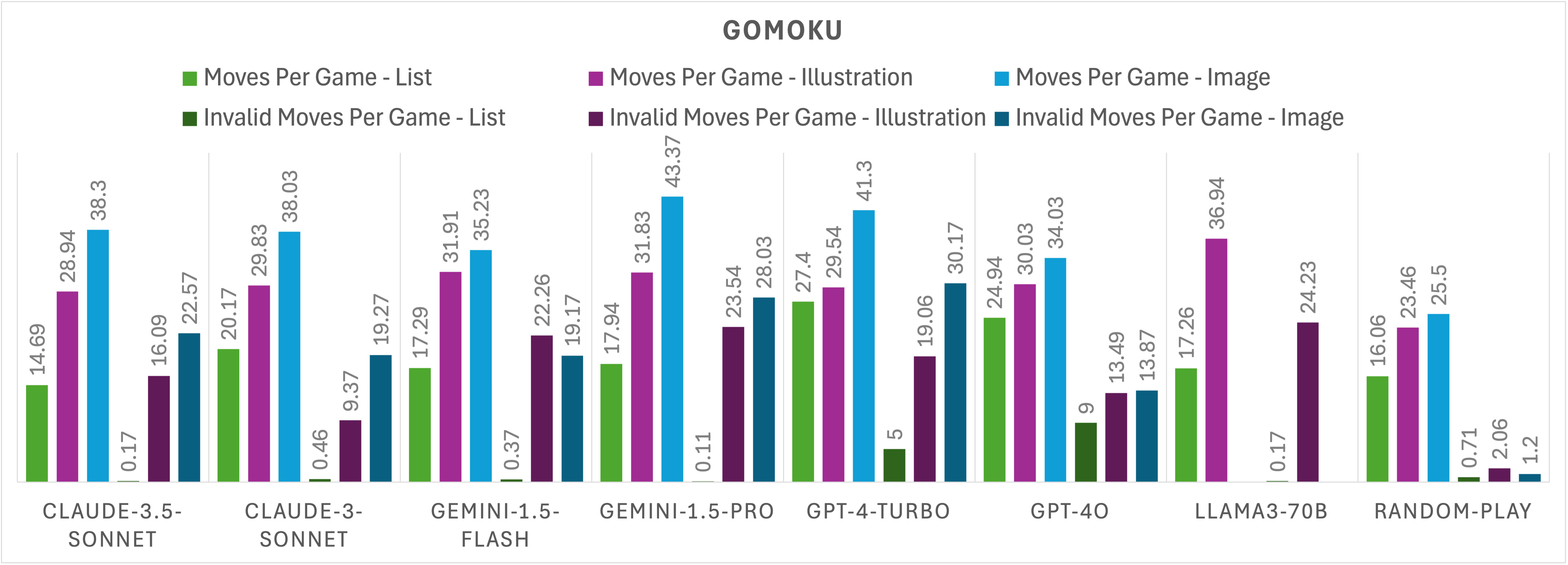}
  \caption{Moves per game and invalid moves (already taken) per game for Gomoku.}
  \label{fig:GomokuInvalidMoves}
\end{figure}

Figure \ref{fig:GomokuInvalidMoves} illustrates the performance of LLMs and a random play strategy in terms of moves per game and invalid moves per game for Gomoku, across list, illustration, and image prompt types. Generally, the number of moves per game and the number of invalid moves per game increases for LLMs with the complexity of the prompt, with the highest moves recorded in the image prompt format. Invalid moves are minimal in list prompts but increase significantly in illustration and image prompts, especially for models like Gemini 1.5 Flash, GPT-4 Turbo, and Llama3-70B. Random play shows relatively fewer invalid moves, as it is less likely to place a move on an already occupied space before the game has progressed significantly in the 15 by 15 grid of Gomoku. This chart highlights the challenges LLMs face in handling more complex and visually demanding prompt formats.

We analyzed the strategic decision-making capabilities of LLMs by counting instances where they missed opportunities to win or block an opponent's win with one move. For example, in Tic-Tac-Toe, if the first player had two of its symbols in a row along with an empty space and did not place its next move in that space to win the game, it was counted as a missed opportunity to win. Similarly, if the second player did not place its next move in the empty space to block the first player from winning after the first player had two symbols in a row, it was recorded as a missed opportunity to block. Our analysis covered 70 games per LLM for the list and illustration prompt types, and 60 games per LLM for the image prompt type.

\begin{figure}[htb]
  \centering
 \includegraphics[width=\textwidth,height=\textheight,keepaspectratio]{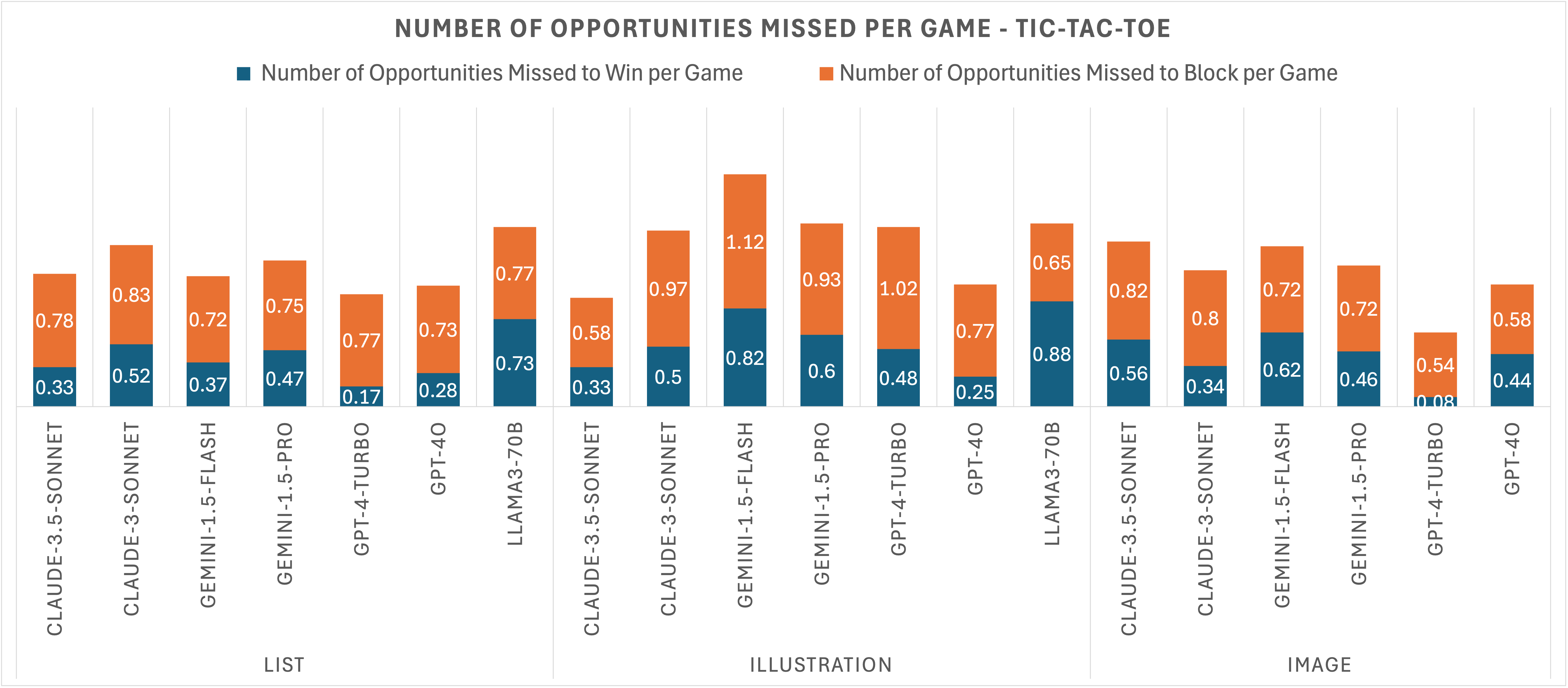}
  \caption{Strategic move opportunities missed per Tic-Tac-Toe game.}
  \label{fig:TicTacToe-OpportunityMissed-PerGame}
\end{figure}

Figure \ref{fig:TicTacToe-OpportunityMissed-PerGame} presents the frequency of missed opportunities to win or avoid a loss per Tic-Tac-Toe game. LLMs generally performed better by missing fewer opportunities in list prompts compared to illustration and image prompts. Claude 3.5 Sonnet showed the fewest missed opportunities to win in the list and illustration prompts, while GPT-4 Turbo showed the fewest in the image prompt. The frequency of missed opportunities to block an opponent's win was generally higher across all prompt types, with Gemini 1.5 Flash facing notable challenges in the illustration prompts.

\begin{figure}[htb]
  \centering
 \includegraphics[width=\textwidth,height=\textheight,keepaspectratio]{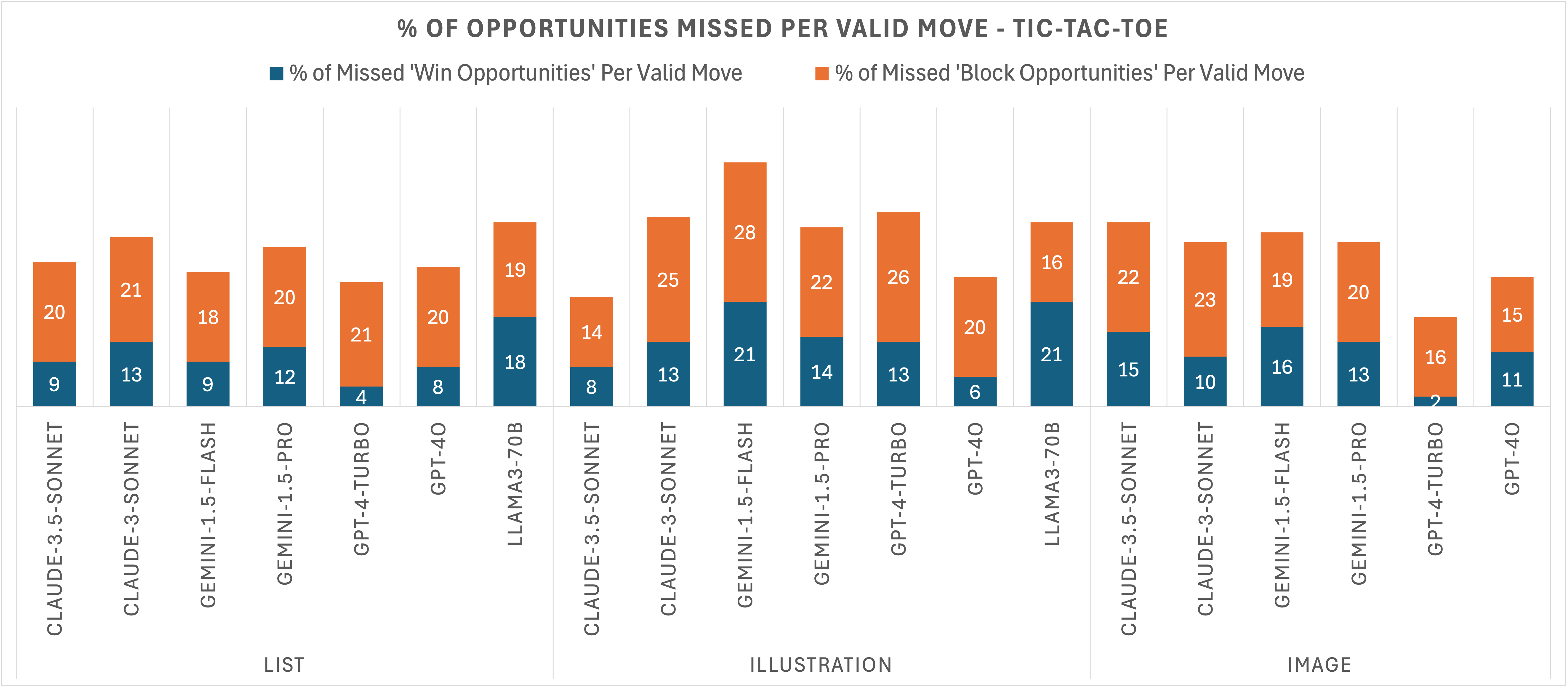}
  \caption{Percentage of strategic move opportunities missed per Tic-Tac-Toe valid move.}
  \label{fig:TicTacToe-OpportunityMissed-PerValidMove}
\end{figure}

In general, missing fewer win and block opportunities per game indicates better performance for an LLM. However, if an LLM makes many invalid moves and gets disqualified without creating any opportunities to win, the number of missed opportunities to win will be zero, falsely suggesting no missed opportunities. To avoid such confusion in the interpretation of results and to further analyze performance, we normalized the missed opportunities by using the number of valid moves and calculated the percentage of opportunities missed per valid move. The chart in Figure \ref{fig:TicTacToe-OpportunityMissed-PerValidMove} shows the percentage of missed win and block opportunities per valid move for various LLMs in Tic-Tac-Toe across three prompt types: list, illustration, and image. The blue bars rep-resent the percentage of missed win opportunities, while the orange bars represent the percentage of missed block opportunities. Generally, LLMs missed fewer opportunities in the list prompts compared to illustration and image prompts. For instance, Claude 3.5 Sonnet had a 9\% missed win opportunity rate and 20\% missed block opportunity rate for list prompts, while it missed 21\% and 28\%, respectively, for illustration prompts. Similarly, GPT-4 Turbo had a notable increase in missed block opportunities for illustration prompts. The trend is consistent across other models, indicating that LLMs face greater challenges in handling visually complex prompts, leading to higher rates of missed strategic opportunities. The chart highlights that while LLMs can identify winning moves, they often struggle more with blocking opponents' winning moves, especially as prompt complexity increases.

\begin{figure}[htb]
  \centering
 \includegraphics[width=\textwidth,height=\textheight,keepaspectratio]{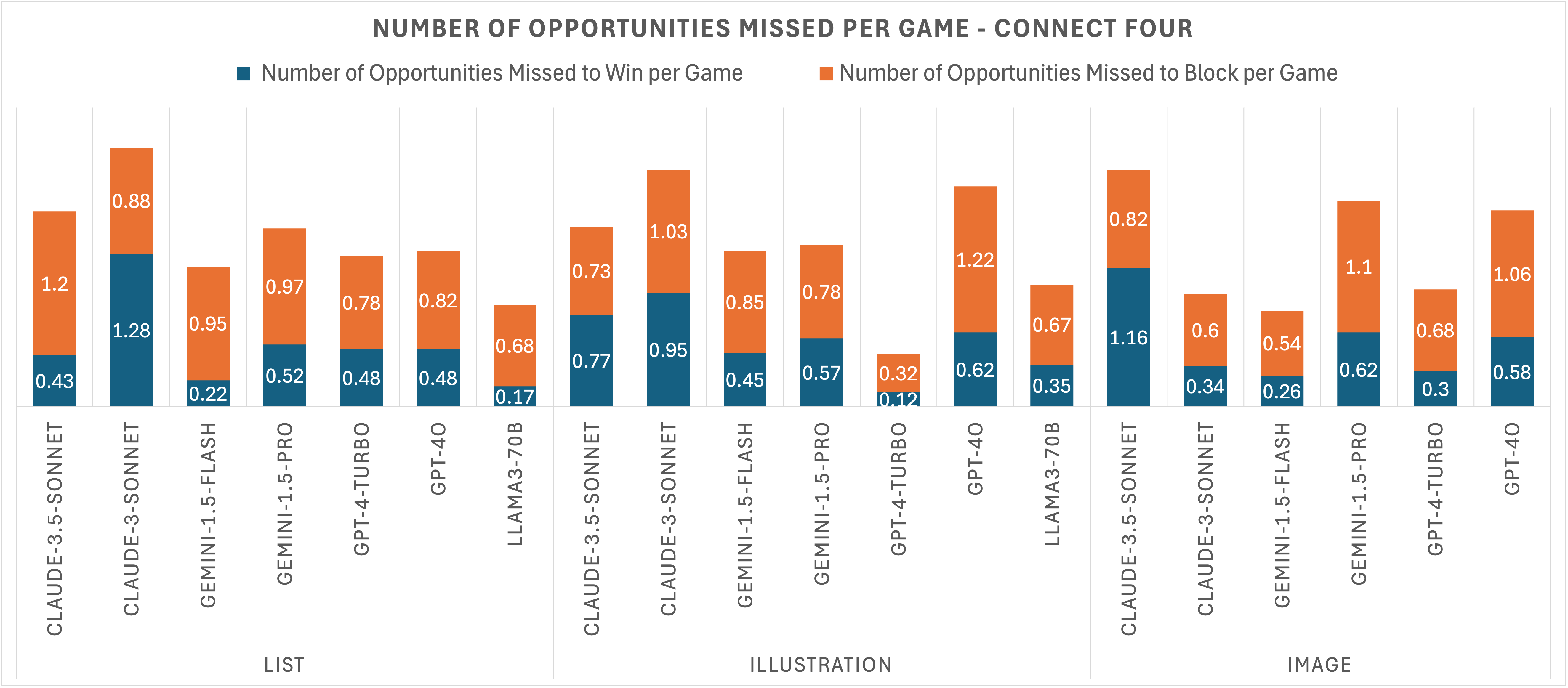}
  \caption{Strategic move opportunities missed per Connect Four game.}
  \label{fig:Connect4-OpportunityMissed-PerGame}
\end{figure}

Figure \ref{fig:Connect4-OpportunityMissed-PerGame} shows the performance of various LLMs in terms of missed opportunities to either win or block the opponent from winning in Connect Four. In Connect Four, if a player had three of its discs in a row (horizontally, vertically, or diagonally) along with an empty slot and did not place its next move in that slot to complete four in a row and win the game, it was counted as a missed opportunity to win. Similarly, if the opponent did not place its next move in the empty slot to block the first player from winning once the first player had achieved three discs in a row, it was recorded as a missed opportunity to block. According to the chart, LLMs tend to miss more opportunities to block than to win. For example, in the list prompt format, Claude 3 Sonnet has a high rate of missed opportunities to block at 1.28 per game, while it misses 0.88 opportunities to win. Similarly, Gemini 1.5 Flash misses 0.95 opportunities to block and 0.22 to win. This trend is consistent across other models and prompt types, indicating a common difficulty among LLMs in anticipating the opponent's winning moves. In the image prompt format, Claude 3.5 Sonnet misses 0.68 opportunities to block and 0.97 to win, indicating challenges in both defensive and offensive strategies. In the illustration prompt, GPT-4o and Claude 3 Sonnet miss the most opportunities compared to other LLMs.

\begin{figure}[htb]
  \centering
 \includegraphics[width=\textwidth,height=\textheight,keepaspectratio]{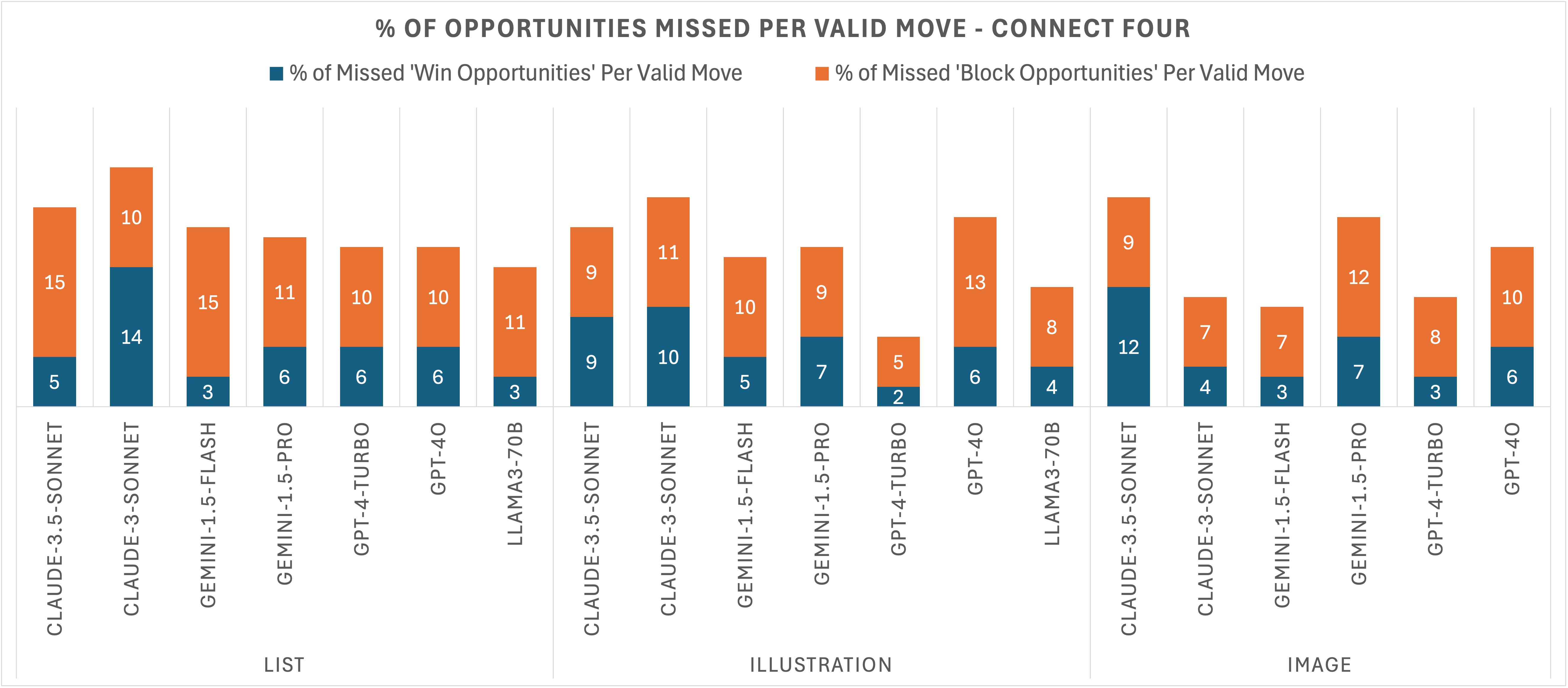}
  \caption{Percentage of strategic move opportunities missed per Connect Four valid move.}
  \label{fig:Connect4-OpportunityMissed-PerValidMove}
\end{figure}

GPT-4 Turbo appears to miss fewer opportunities than other LLMs in the illustration prompt format. However, a careful analysis of the previous chart in Figure \ref{fig:Connect4Illustration} reveals that GPT-4 Turbo had a disqualification rate of 82.66\% as the first player and 68.57\% as the second player out of a total of 70 games using the illustration prompt in Connect Four. These rates are significantly higher than those of other LLMs. Further review of Figure \ref{fig:Connect4InvalidMoves} shows that GPT-4 Turbo averaged 13.49 invalid moves per game out of 23.43 moves, which is again significantly more than other LLMs. While GPT-4 Turbo seems to have missed fewer opportunities, this may result from not having many valid moves. To gain better insight into the LLMs' strategic capabilities in utilizing opportunities, we focused on valid moves and analyzed the percentage of missed win and block opportunities per valid move. Figure \ref{fig:Connect4-OpportunityMissed-PerValidMove} shows the results of this analysis. For list prompts, Claude 3 Sonnet missed the most block opportunities. In illustration prompts, Llama3-70B and GPT-4 Turbo performed better. In image prompts, missed block opportunities were notably higher for Claude 3.5 Sonnet and Gemini 1.5 Pro.

\begin{figure}[htb]
  \centering
 \includegraphics[width=\textwidth,height=\textheight,keepaspectratio]{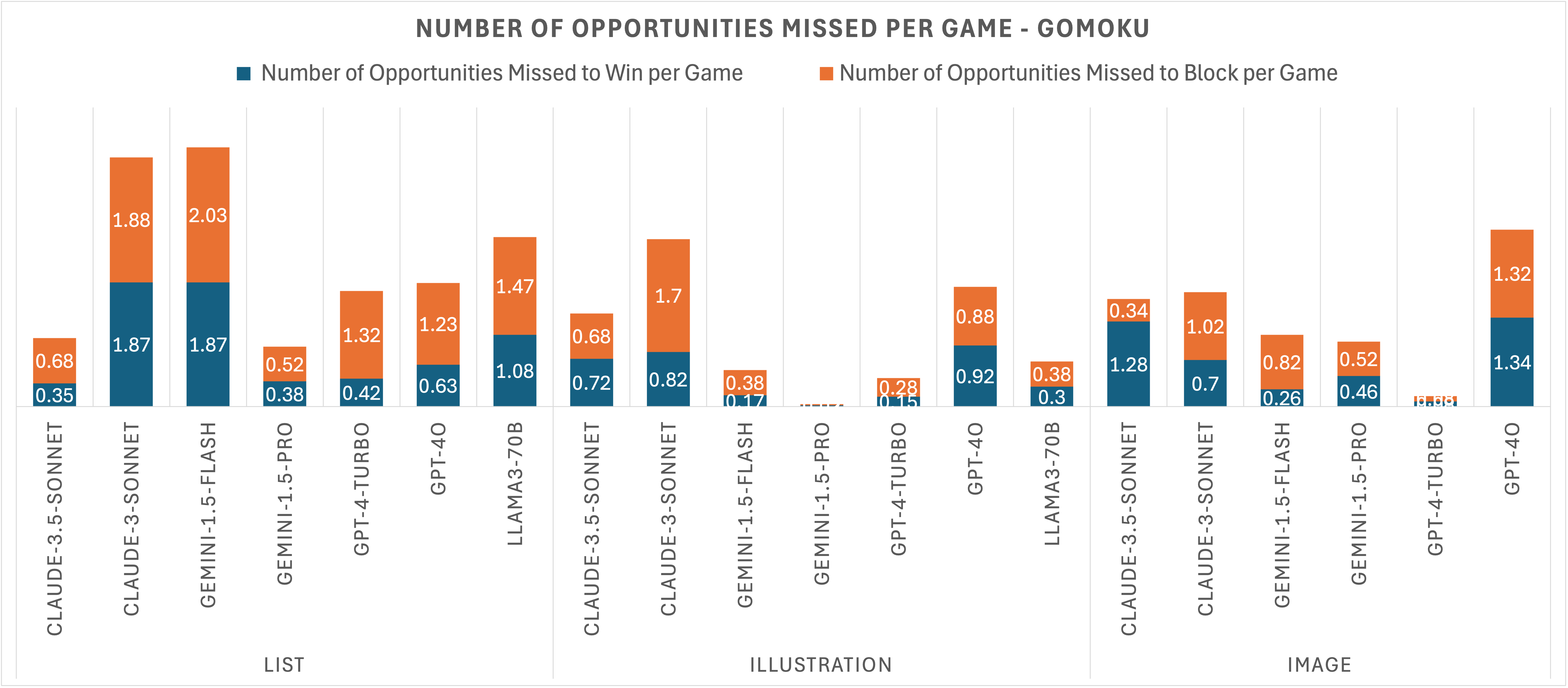}
  \caption{Strategic move opportunities missed per Gomoku game.}
  \label{fig:Gomoku-OpportunityMissed-PerGame}
\end{figure}

According to the chart in Figure \ref{fig:Gomoku-OpportunityMissed-PerGame}, LLMs generally miss more opportunities to win than to block in the Gomoku game. In Gomoku, if a player has four stones in a row (horizontally, vertically, or diagonally) and an empty space but does not place its next move in that space to complete five in a row, it is counted as a missed opportunity to win. Similarly, if the opponent does not place its next move in the empty space to block the first player from winning once the first player has four stones in a row, it is recorded as a missed opportunity to block. In the list prompt format, both Claude 3 Sonnet and Gemini 1.5 Flash miss approximately 1.87 opportunities to win per game, while their missed opportunities to block are slightly lower. In the illustration prompt format, Gemini 1.5 Flash misses 1.7 opportunities to win and 0.82 to block. This trend indicates that while LLMs can often identify opportunities to block, they struggle more with recognizing and capitalizing on winning opportunities. In the image prompt format, GPT-4o exhibits the highest number of missed opportunities, with 1.34 missed chances to win and 1.32 to block per game. This highlights significant challenges for LLMs in processing and acting upon image-based inputs. Overall, the chart emphasizes that while LLMs are reasonably adept at blocking the opponent's winning moves, they face greater difficulty in identifying and seizing their own winning opportunities, especially in more complex and visually demanding formats.

\begin{figure}[htb]
  \centering
 \includegraphics[width=\textwidth,height=\textheight,keepaspectratio]{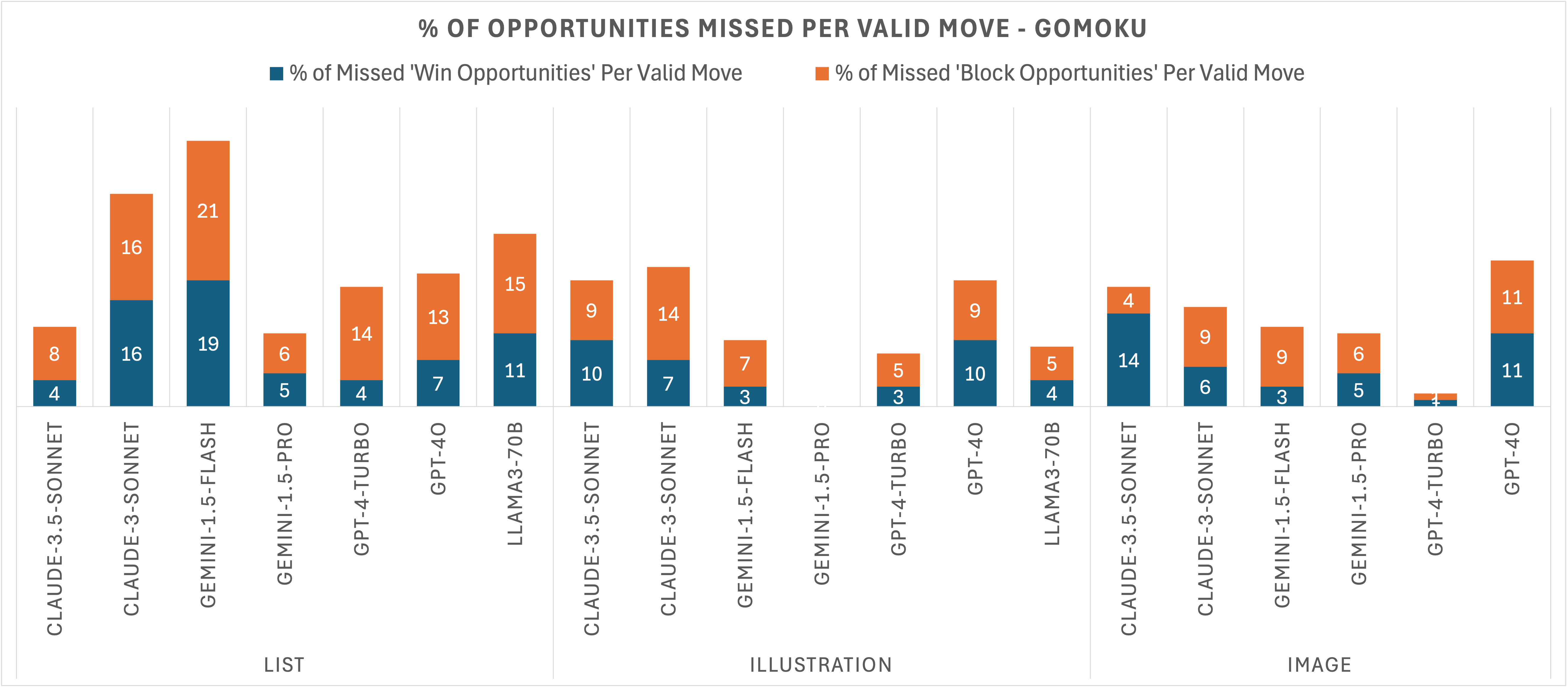}
  \caption{Percentage of strategic move opportunities missed per Gomoku valid move.}
  \label{fig:Gomoku-OpportunityMissed-PerValidMove}
\end{figure}

In Figure \ref{fig:Gomoku-OpportunityMissed-PerValidMove}, we further analyzed the missed opportunities by normalizing them per valid move, as we did in Figure \ref{fig:TicTacToe-OpportunityMissed-PerValidMove} for Tic-Tac-Toe and Figure \ref{fig:Connect4-OpportunityMissed-PerValidMove} for Connect Four. Figure \ref{fig:Gomoku-OpportunityMissed-PerValidMove} displays the percentage of missed win and block opportunities per valid move for various LLMs in Gomoku across three prompt types: list, illustration, and image. Generally, the LLMs show a higher percentage of missed block opportunities compared to win opportunities across all prompt types. In list prompts, Gemini 1.5 Flash missed the most block opportunities. Gemini 1.5 Pro performed the best with no missed opportunities for illustration prompts. In image prompts, GPT-4 Turbo performed the best in minimizing missed opportunities.

Utilizing the Game Simulation web app to generate data for new LLMs or the open-access data available on GitHub for the LLMs assessed here, further analysis can be conducted. For instance, analyzing the creation of winning opportunities can provide additional insights. This can be achieved by counting instances where the LLM creates potential winning opportunities (i.e., aligning two moves for Tic-Tac-Toe, three moves for Connect Four, and four moves for Gomoku). Such analysis can reveal that a high number of created opportunities may indicate proactive strategic thinking by the LLMs. Contributions to the repository with suggestions and new evaluation metrics are encouraged to enhance the assessment of LLM capabilities in grid-based games.

\begin{figure}[htb]
  \centering
 \includegraphics[width=\textwidth,height=\textheight,keepaspectratio]{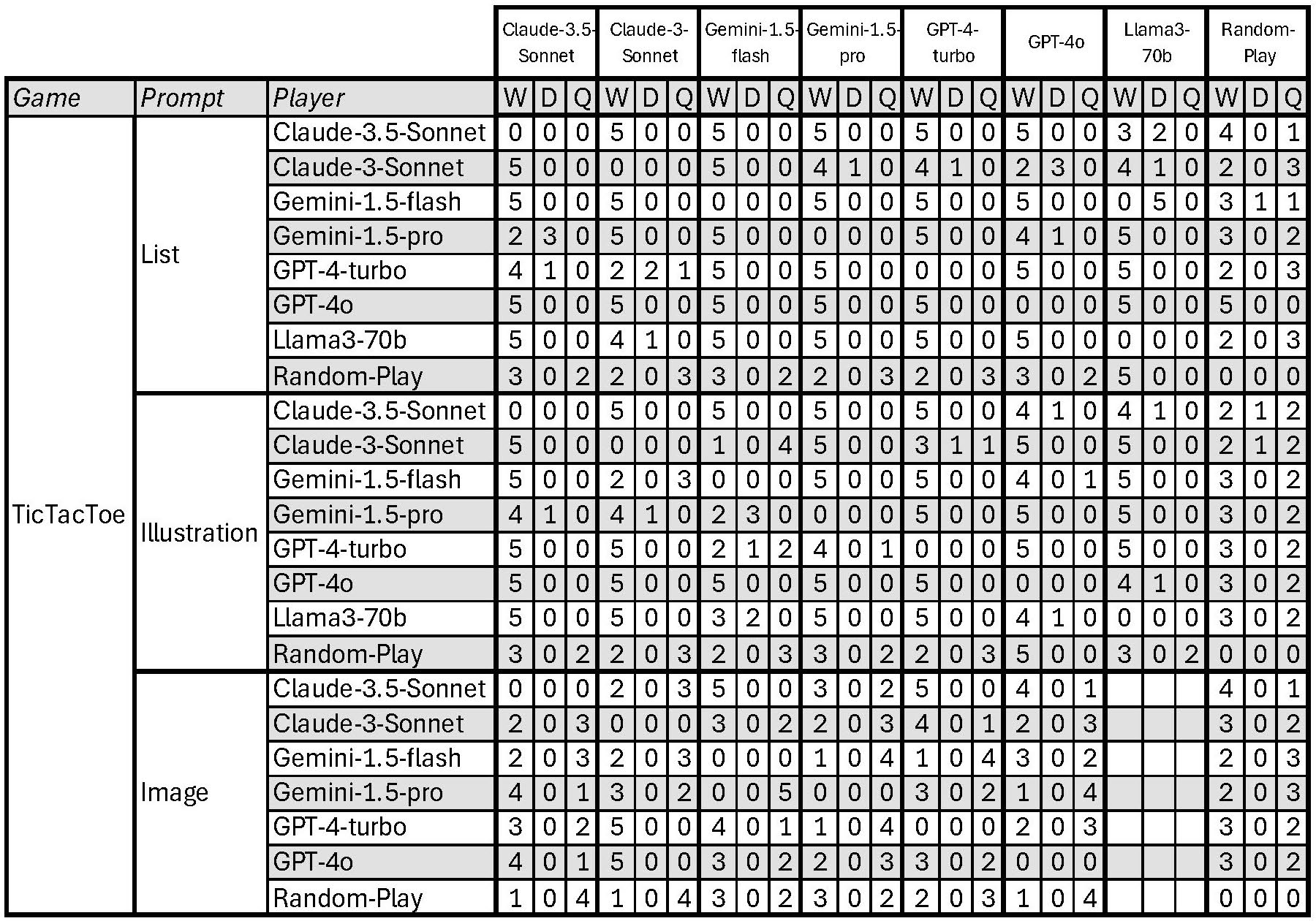}
  \caption{The results matrix showing the outcomes of matches between seven LLMs and a random play generator software across different games for Tic-Tac-Toe.}
  \label{fig:TicTacToe-ResultsMatrix}
\end{figure}

\begin{figure}[htb]
  \centering
 \includegraphics[width=\textwidth,height=\textheight,keepaspectratio]{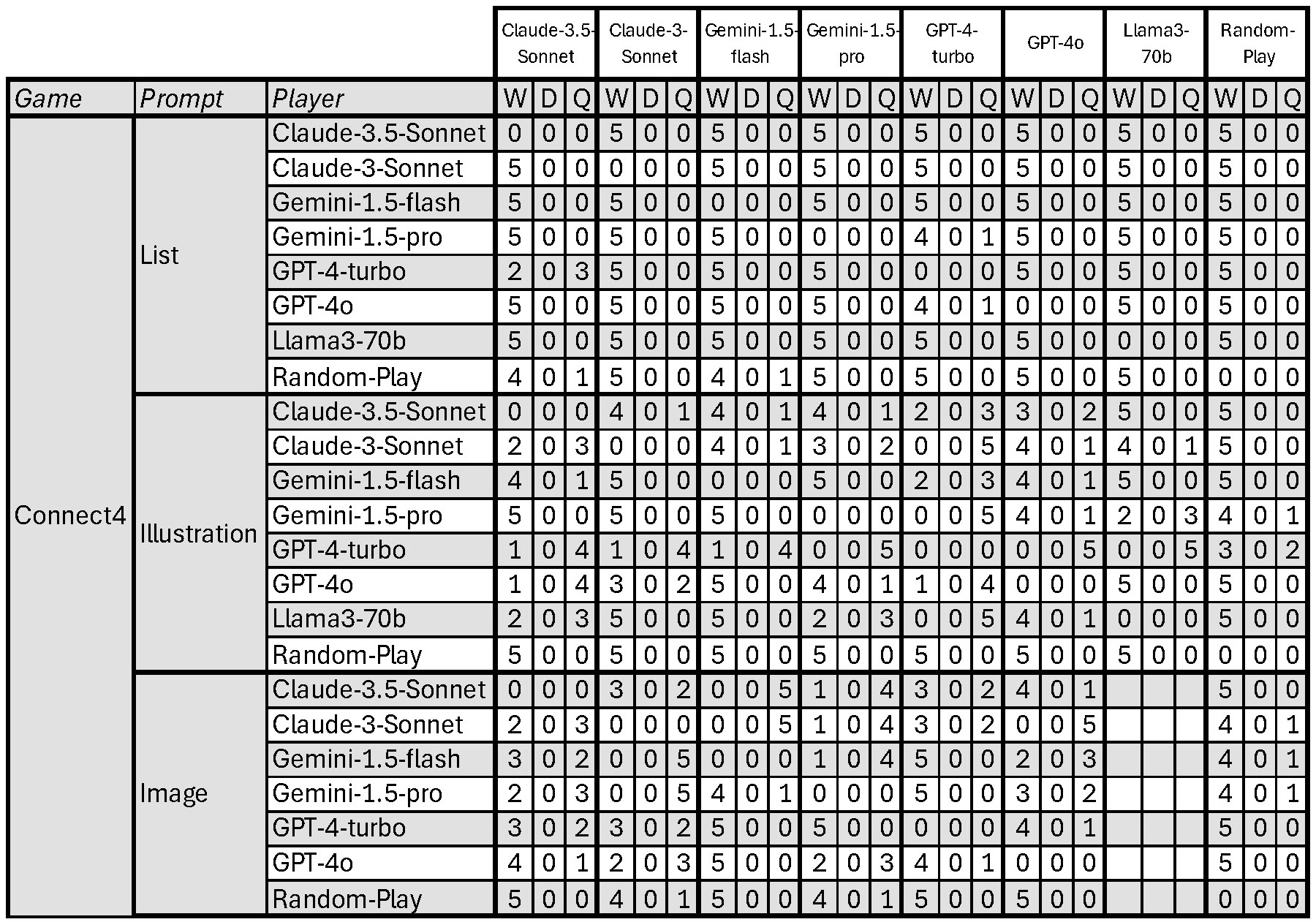}
  \caption{The results matrix showing the outcomes of matches between seven LLMs and a random play generator software across different games for Connect Four.}
  \label{fig:Connect4-ResultsMatrix}
\end{figure}

\begin{figure}[htb]
  \centering
 \includegraphics[width=\textwidth,height=\textheight,keepaspectratio]{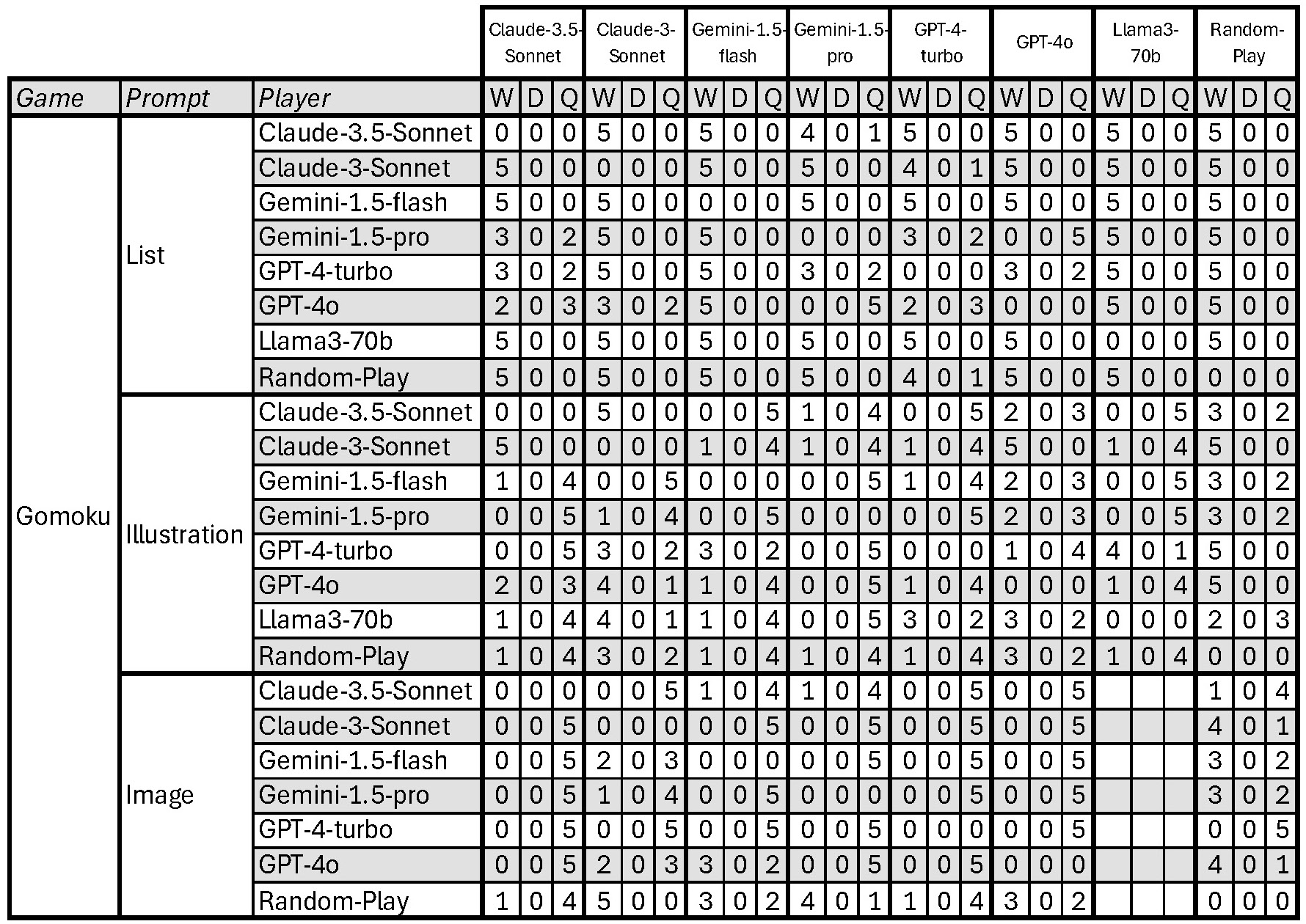}
  \caption{The results matrix showing the outcomes of matches between seven LLMs and a random play generator software across different games for Gomoku.}
  \label{fig:Gomoku-ResultsMatrix}
\end{figure}

Figures \ref{fig:TicTacToe-ResultsMatrix}, \ref{fig:Connect4-ResultsMatrix} and \ref{fig:Gomoku-ResultsMatrix} summarize the outcomes of 2,310 games played between seven LLMs and a random play generator across different prompt types (list, illustration, image) for the games Tic-Tac-Toe, Connect Four, and Gomoku, respectively. As shown in these results matrices, the LLMs demonstrated varying degrees of success across different games and prompt types. Models like Claude 3.5 Sonnet, GPT-4o, and Gemini 1.5 Pro showed strong performance in simpler formats but struggled with more complex prompts. Random play consistently had the highest number of invalid moves, highlighting its lack of strategy. The data suggests that while some LLMs handle simpler game prompts well, more complex formats reveal significant challenges in their decision-making processes. In the result matrices, the W column shows the total number of games the LLM won as the first and second player, D indicates draws, and Q shows the total number of games the LLM was disqualified as the first and second player. Each LLM played 5 games with each corresponding opponent for each game and prompt type combination, totaling 35 games per player for list and illustration prompts, and 30 games per player for the image prompt type. The lower game count for the image prompt type is because Llama3-70B cannot accept images and therefore was not used with the image prompt type. The result matrix is also hosted on the project's GitHub page.

\begin{figure}[htb]
  \centering
 \includegraphics[width=\textwidth,height=\textheight,keepaspectratio]{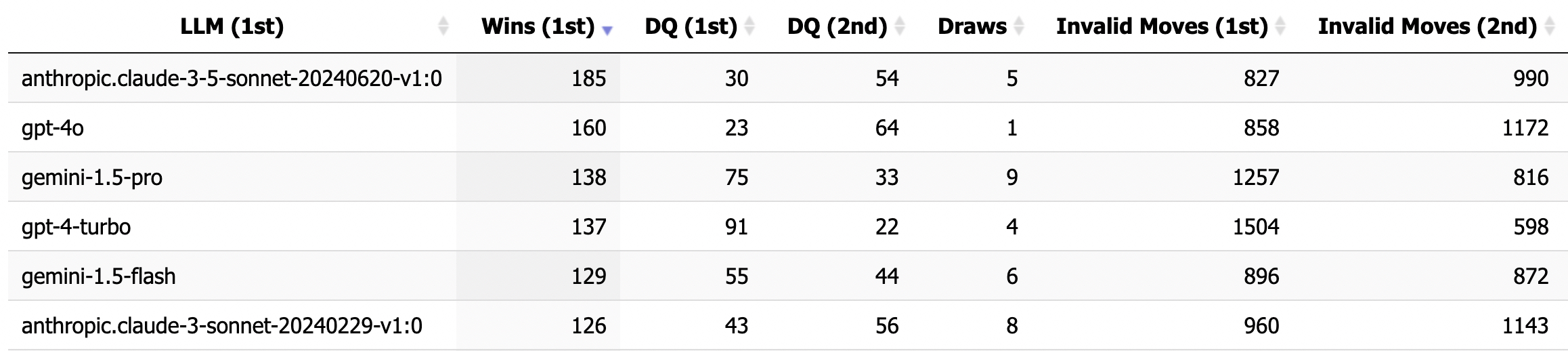}
  \caption{A snapshot from the leaderboard showing aggregated results.}
  \label{fig:Leaderboard-Snapshot}
\end{figure}

Figure \ref{fig:Leaderboard-Snapshot} displays a portion of the leaderboard page of the LLM Game Benchmark, summarizing the results of games played between various LLMs and a random play generator. Key metrics on the leaderboard include win ratios, number of wins, dis-qualifications, invalid moves, and total moves for both the first and second players. Users can filter games by type, prompt type, and LLM players, and sort results by clicking on any of the column headers. Functionality to aggregate results by game type, prompt type, and LLM player is provided on the leaderboard page. The initial data includes results from Claude 3.5 Sonnet, Claude 3 Sonnet, Gemini 1.5 Flash, Gemini 1.5 Pro, GPT-4 Turbo, GPT-4o and Random-Play. Each LLM played against each other LLM five times for each game and prompt type combination. New game result submissions are welcome and can be generated using the game simulation web software. The leaderboard web page can be accessed on the benchmark's GitHub page \footnote{LLM Grid-Based Game Leaderboard: \url{https://research-outcome.github.io/LLM-Game-Benchmark/leaderboard/}} \cite{GitHub}. The data used to fill in the leaderboard page can be downloaded in JSON format.

\section{Discussion}
This study provides a comprehensive evaluation of seven LLMs, Claude 3.5 Sonnet, Claude 3 Sonnet, Gemini 1.5 Flash, Gemini 1.5 Pro, GPT-4 Turbo, GPT-4o, and Llama3-70B, alongside a random play generator across three games (Tic-Tac-Toe, Connect Four, and Gomoku) and three prompt types (list, illustration, image). Each LLM played five games per game and prompt type against each opponent, resulting in a total of 2,310 games for analysis.

The performance of the LLMs varied significantly across different games and prompt types. Simpler games like Tic-Tac-Toe experienced fewer invalid moves and disqualifications compared to more complex games like Connect Four and Gomoku, highlighting the models' varying capacities to handle increasing game complexity. LLMs performed better with list prompts for Tic-Tac-Toe and Connect Four, while more complex prompt formats, particularly illustration and image prompts, revealed challenges in strategic decision-making. These formats led to higher disqualification rates and missed strategic opportunities, indicating difficulties in interpreting visual data and maintaining consistency.

The random play strategy consistently recorded the highest number of losses and invalid moves, serving as a useful baseline for gauging LLM performance. The stark contrast between random play and the LLMs underscores the models' capacity for strategic decision-making, though there remains room for improvement in handling complex and visual data.

In Tic-Tac-Toe, LLMs demonstrated strong performance with list prompts, exhibiting minimal invalid moves, indicating a good understanding of the game's basic rules. Performance declined with illustration prompts, as some LLMs showed increased invalid moves, suggesting challenges in interpreting visual prompts. The most significant decline was observed with image prompts, where many LLMs displayed a high number of in-valid moves, highlighting difficulties in processing and responding to image-based prompts.

In Connect Four, LLMs generally performed well with list prompts, although some showed increased invalid moves compared to Tic-Tac-Toe, reflecting the higher complexity of Connect Four. A significant increase in invalid moves and disqualifications was observed with both illustration and image prompts, indicating substantial challenges in visual interpretation and decision-making in a more complex game context.

For Gomoku, performance was mixed across all prompt types, with a notable increase in invalid moves and disqualifications. This game, being more complex and less common, likely posed significant interpretative and decision-making challenges for the LLMs.

Different prompt types had a notable impact on the performance of the LLMs. List prompts were generally well-handled by all LLMs, suggesting that textual representation of game states is within their current capabilities. Illustration prompts posed moderate challenges, as reflected in the increased number of invalid moves, indicating that graphical representations are harder for LLMs to interpret. Image prompts were the most challenging, with the highest number of invalid moves, highlighting a significant area for improvement in LLMs' ability to process and act on image-based inputs.

Invalid moves analysis revealed no out-of-bounds errors in any games, unlike in our previous study. This improvement is likely due to updated prompts that clearly define the range of possible column and row values. Invalid format errors were made only by GPT-4 Turbo and Gemini 1.5 Flash, mostly due to hallucinated tag names in the JSON format. GPT-4 Turbo made several invalid format errors in Tic-Tac-Toe and Gomoku games for list and illustration prompt types, while Gemini 1.5 Flash made several errors during Gomoku games for illustration and image prompts. A significant percentage of the invalid moves were due to moving to an already occupied space, as shown in Figures \ref{fig:TicTacToeInvalidMoves}, \ref{fig:Connect4InvalidMoves}, and \ref{fig:GomokuInvalidMoves}.

The analysis of missed strategic opportunities highlights the variability in LLMs' decision-making processes. Models like Claude 3.5 Sonnet and GPT-4 Turbo showed fewer missed opportunities in list prompts, suggesting a better grasp of straightforward game mechanics. However, the higher frequency of missed opportunities in illustration and image prompts indicates that LLMs struggle with interpreting and acting upon visual data.

While the study uses games as a benchmark, the findings have broader implications for LLM applications in fields such as robotics, autonomous systems, and interactive AI. Improving LLMs' strategic thinking and decision-making abilities can enhance their performance in various real-world tasks requiring similar cognitive skills.

The extensible nature of the benchmark, with its modular code and open invitation for community contributions, represents a significant step towards collaborative LLM research. Encouraging researchers to add new games and share their results can lead to a more dynamic and comprehensive evaluation framework. Future work can include a broader range of games and tasks to evaluate LLMs across different strategic environments.

\section{Limitations and Future Directions}
The study's methodology primarily focuses on grid-based games, which, while useful, may not fully capture the breadth of real-world strategic interactions. Future benchmarks should incorporate a wider variety of game types, including those with more complex rules and longer-term strategic planning, to provide a more comprehensive assessment of LLM capabilities. Designing new, custom, purpose-built games to test specific aspects of LLM capabilities, such as adapting to unusual rules, would enhance benchmarking effectiveness and prevent the possibility of LLMs becoming familiar with the game, even if they were not specifically trained for it.

The simplicity of the games used in this benchmark facilitates basic evaluation but may not challenge LLMs' strategic capabilities as much as more complex games like chess or Go might. The fact that current LLMs have not mastered even these simple games provides valuable insights into their capabilities and limitations. Expanding the evaluation to larger grids, such as 4 × 4 or 5 × 5 for Tic-Tac-Toe or 19 × 19 for Gomoku, could present additional challenges and provide a clearer indicator of LLM performance.

Relying on predefined prompts to guide LLMs' moves may not adequately capture their potential for independent strategic thinking or their ability to respond to changing game states. Although we updated prompts dynamically to warn LLMs of invalid moves, further techniques, such as providing all previous invalid moves, could be explored to reduce invalid move numbers and disqualifications.

This study tested LLMs using structured prompts. Future research should investigate how these prompts influence LLM performance and how variations in prompt structure might affect their understanding of game states and subsequent moves. Such insights could help optimize LLMs for more complex and varied applications.

The evaluation metrics used in this study revealed a wide range of LLM capabilities. While these metrics provide a good indication of performance, they may not fully capture the strategic complexity of the models. Further analysis of the moves—drawn from the JSON and PNG files—could offer a more detailed assessment of game progress over time. The new analysis can be conducted using the Game Simulation web app to generate data for new LLMs or the open-access data on GitHub. For example, evaluating the creation of winning opportunities, such as aligning moves for Tic-Tac-Toe, Connect Four, and Gomoku, can provide insights into proactive strategic thinking by the LLMs. The authors encourage and welcome contributions to the repository in the form of suggestions and implementations of new metrics and methods to evaluate the capabilities of LLMs.

Focusing on a select group of LLMs might not capture the full diversity of strategic approaches across available models, highlighting the importance of including a broader array in future research. The rapidly expanding landscape of LLMs, with new models and improved versions emerging frequently, necessitates continuous updates to benchmarks. We welcome submissions of other and new LLMs using the open-source game simulation software.

Future work could explore several promising directions to extend research and deepen our understanding of LLM capabilities in strategic games and beyond. Multi-agent collaboration scenarios, where multiple LLMs work together against a common opponent or compete in teams, could assess their abilities in coordination, cooperation, and competitive strategy. Comparing newer versions of LLMs against those tested in this study could track progress and improvements in AI strategic gaming capabilities over time.

This study suggests several avenues for future research and development. Firstly, improving LLMs' abilities to interpret and act on visual data is crucial, as evidenced by high invalid move rates in illustration and image prompts. Enhancing visual processing capabilities could significantly boost overall performance and utility. Secondly, further research is needed to enhance LLMs' decision-making processes in more complex environments.

\section{Conclusion}
This study introduces a novel and extensible benchmark for LLMs through grid-based games such as Tic-Tac-Toe, Connect Four, and Gomoku. The open-source game simulation code, available on GitHub, enables LLMs to compete and generates data files in JSON, CSV, TXT, and PNG formats for leaderboard rankings and further analysis. We present the results of games among leading LLMs, including Claude 3.5 Sonnet and Claude 3 Sonnet by Anthropic, Gemini 1.5 Pro and Gemini 1.5 Flash by Google, GPT-4 Turbo and GPT-4o by OpenAI, and Llama3-70B by Meta, and encourage submissions from other LLMs. By analyzing the performance of these models over 2,310 games, we observed significant variations in their capabilities, particularly highlighting their struggles with complex and visually-based prompt formats. Comparisons with a random play generator underscore the LLMs' superior yet still developing capacity for strategic decision-making.

The study reveals that while LLMs perform relatively well in simpler formats, such as list prompts for Tic-Tac-Toe and Connect Four, their performance declines with more complex prompts, especially those involving illustrations and images. This trend indicates the current limitations in LLMs' ability to interpret and act on visual data and manage increased game complexity. Additionally, the models showed a tendency to make invalid moves when faced with more complex prompts, underscoring the need for improved strategic decision-making processes.

Several areas for further investigation could be explored, such as expanding the types and complexity of games used for evaluation, testing more sophisticated prompt engineering techniques, and delving deeper into the effects of different prompt structures.

The findings of this study have broader implications beyond gaming, suggesting that advancements in LLMs' strategic thinking and decision-making abilities could enhance their application in fields such as robotics, autonomous systems, and interactive AI. Furthermore, the modular and open-source nature of the benchmarking framework encourages community contributions, which can lead to a more dynamic and comprehensive evaluation of LLM capabilities.

In conclusion, while the current evaluation highlights both the strengths and limitations of LLMs, it also points to the need for ongoing research to enhance their ability to process complex and visual data, improve decision-making processes, and develop more sophisticated benchmarking tools. This continuous development will ultimately broaden the applicability and effectiveness of LLMs in various real-world tasks.

\section*{Acknowledgments}
This study was partially supported by Florida Polytechnic University with grant number GR-24SUMR-OT.

\bibliographystyle{unsrt}  
\bibliography{LLM-Game-Benchmark}

\end{document}